\documentclass{article}

\usepackage{PRIMEarxiv}

\usepackage[utf8]{inputenc} 
\usepackage[T1]{fontenc}    
\usepackage[hidelinks]{hyperref}       
\usepackage{url}            
\usepackage{booktabs}       
\usepackage{amsfonts}       
\usepackage{nicefrac}       
\usepackage{microtype}      
\usepackage{mathtools}      
\usepackage{fancyhdr}       
\usepackage{graphicx}       
\usepackage{orcidlink}      
\usepackage{wrapfig}        
\usepackage{multicol}       
\usepackage{color,soul}
\usepackage{subcaption}

\usepackage[font=footnotesize,labelfont=bf]{caption}

\title{VoloGAN: Adversarial Domain Adaptation for Synthetic Depth Data}

\author{
  Sascha Kirch \orcidlink{0000-0002-5578-7555} \\
  UNED - Universidad Nacional de Educación a Distancia \\
  Madrid, Spain \\
  \href{mailto:skirch1@alumno.uned.es}{\texttt{skirch1@alumno.uned.es}} \\
   \And
  Sergio Arnaldo \\
  Volograms Ltd. \\
  Dublin, Ireland \\
  \href{mailto:sergio@volograms.com}{\texttt{sergio@volograms.com}} \\
  \AND
  Sergio Martín \orcidlink{0000-0002-4118-0234} \\
  UNED - Universidad Nacional de Educación a Distancia \\
  Madrid, Spain \\
  \href{mailto:smartin@ieec.uned.es}{\texttt{smartin@ieec.uned.es}} \\
  \And
  Rafael Pagés \orcidlink{0000-0002-5691-9580} \\
  Volograms Ltd. \\
  Dublin, Ireland \\
  \href{mailto:rafa@volograms.com}{\texttt{rafa@volograms.com}} \\
}

\begin{document}
\maketitle

\begin{abstract}
We present VoloGAN, an adversarial domain adaptation network that translates synthetic RGB-D images of a high-quality 3D model of a person, into RGB-D images that could be generated with a consumer depth sensor. This system is especially useful to generate high amount training data for single-view 3D reconstruction algorithms replicating the real-world capture conditions, being able to imitate the style of different sensor types, for the same high-end 3D model database. The network uses a CycleGAN framework with a U-Net architecture for the generator and a discriminator inspired by SIV-GAN. We use different optimizers and learning rate schedules to train the generator and the discriminator. We further construct a loss function that considers image channels individually and, among other metrics, evaluates the structural similarity. We demonstrate that CycleGANs can be used to apply adversarial domain adaptation of synthetic 3D data to train a volumetric video generator model having only few training samples. 
\end{abstract}


\section{Introduction}
\label{sec:introduction}

With the rise of immersive technologies, such as Virtual and Augmented Reality, and more recently, the Metaverse, we are seeing a new wave of opportunities in the shape of applications that go from navigation~\cite{ruano2017augmented}, to communications~\cite{orts2016holoportation}, to many more. In this context, bringing real humans to these immersive platforms in an easy and affordable way remains a challenge. High-end volumetric video techniques, such as the one proposed by Collet et al.~\cite{collet2015high} or Guo et al.~\cite{guo2019relightables} use very complicated camera setups that include different types of sensors, professional lighting and a huge amount of data to process. Other more affordable techniques, such as the one proposed by Pag\'es et al.~\cite{pages_affordable_2018}, approach the problem assuming there will be a reduced number of cameras in an uncontrolled environment (outdoors, handheld devices), but still require multiple recording devices. Newer techniques, such as PIFu~\cite{saito_pifu_2019, saito_pifuhd_2020}, ARCH++~\cite{he2021arch++} or Volu~\cite{volu} propose using a single input image to generate a full 3D model of a person, including unseen areas, like the back of the captured subject. 

These single-view techniques use deep learning approaches that need to be trained with thousands of high-quality 3D models, normally captured with high-end 3D scanners, which are rendered from different viewpoints generating a very large synthetic RGB database. This means that to improve the reconstruction with additional inputs, we need to be able to add that information to the training data. For example, if we wanted to include a depth sensor as an additional layer to acquire more details or a real-world scale to the result obtained with the single-view approach, we would need to re-train our algorithm with the newly acquired data. However, if we would render the depth data as we did with the RGB data, we would obtain a quasi-perfect depth image, which is very different from what a consumer depth sensor would record: a depth image with significant temporal noise (increasing with distance), artifacts around discontinuity edges, etc. Training our network with this quasi-perfect information would result in 3D models that contain all the artifacts produced by the depth sensor. One solution would be to capture your 3D model dataset using the same depth sensor that is going to be used in deployment, however, this means that you would need to capture a new dataset for each specific depth sensor you are using, significantly increasing the complexity of the task and reducing the versatility of the dataset.

In this work, we propose a deep learning based framework that uses image domain adaptation to generate synthetic depth images replicating the style of a consumer depth sensor, which can be adapted to different depth sensor technologies, and can be used to generate set of images to train a single-view 3D reconstruction system, focusing on real humans. 

Starting with a dataset of high-quality human 3D models~\cite{volograms2021}, we generate a set of RGB-D images, which we will refer to as ``synthetic data'', that follow a distribution $p_{synthetic}$. Similarly, the data that is generated by our system will be referred to as ``target data'', that follows the target distribution $p_{target}$. And, to learn the mapping from $p_{synthetic}$ to $p_{target}$, we propose the VoloGAN framework.

The main contribution of this paper are:
\begin{enumerate}
  \item A CycleGAN based framework, the VoloGAN, that performs the joint domain adaptation of images and their respective depth.
  \item The proposal of a channel wise loss function, where each channel can be weighted independently to mitigate the impact of channel pollution caused by the early fusion approach of the image and depth channels.
  \item The implementation of a gated self-attention block where the gate is controlled by a learnable scalar.
  \item The evaluation of our method using a dataset containing monocular images and depth maps of humans recorded with a consumer depth sensor and camera.
\end{enumerate}
\section{Related Work}
\label{sec:related_work}

\noindent\textbf{Domain adaptation.} We have observed different approaches of incorporating knowledge of different domains into a model. We cluster them into discriminative approaches and generative approaches. The discriminative approaches aim to increase for example the classification performance on a target dataset (usually few samples available) by adding data from a source domain (usually many samples available).\cite{ganin_unsupervised_2015} has shown an unsupervised approach for images. Recent studies extend this approach also to video data \cite{long_video_2022} and to acoustic data \cite{abeser_towards_2022}. Generative approaches focus on generating new data in the target domain from a given source domain. This translation can either be done between data of a shared modality (e.g. image-to-image \cite{xie_self-supervised_2020}, video-to-video \cite{chen_generative_2020}  or virtual-to-real \cite{guo_gan-based_2020}) or between data of different modalities (e.g. segmentation mask to image/video \cite{isola_image--image_2018},\cite{wang_video--video_2018}, video-to-sound \cite{athanasiadis_audiovisual_2020} or image to depth point cloud \cite{hossain_efficient_2022}). We further focus on the later approach, i.e. generating synthetic data by translating data from a source to a target domain, since we believe that it is beneficial to train on data that follows the same distribution as the data available in the target application.

\noindent\textbf{Domain adaptation using GANs}. Generative adversarial networks (GANs) \cite{goodfellow_generative_2014} have demonstrated how high quality images can be generated when training multiple models in an adversarial manner \cite{donahue_adversarial_2017},\cite{wang_high-resolution_2018},\cite{karras_style-based_2019},\cite{karras_analyzing_2020},\cite{karras_alias-free_2021},\cite{sauer_stylegan-xl_2022}. These models generate images by sampling from a low dimensional latent space. The task of domain adaptation is a translation task, meaning input and output dimensions are equal. Following an encoder-decoder architecture and a supervised training paradigm, \cite{isola_image--image_2018} as well as \cite{park_semantic_2019} demonstrated how to translate between two different modalities (e.g. segmentation mask to RGB image) or between domains (e.g. turning a scene in an image from night-time to day-time). \cite{wang_high-resolution_2018} builds upon early models to process high resolution images, which require more compute resources. GANs have been proven effective to perform at supervised image-to-image translation where a ground truth sample of the target image is available. However, in some cases such a ground truth sample is not available. In case of VoloGAN, no sample from a person (with the same pose, appearance, etc.) is available in both domains. 
The absence of paired training samples is addressed in \cite{zhu_unpaired_2020} by the CycleGAN framework that introduces a second generator and a second discriminator to the standard GAN architecture accompanied with a more sophisticated loss function to compensate for the missing ground truth data. Many papers have built upon the CycleGAN framework to adapt it to their needs. Among many contributions, these adaptations aim for example to strengthen the discriminator \cite{sushko_generating_2021},\cite{schonfeld_u-net_2021}, to regularize the model by additional loss terms  such as a channel-wise loss \cite{tang_gesturegan_2019} or structural similarity loss \cite{tang_dual_2019}, normalization techniques such as spectral normalization \cite{miyato_spectral_2018} and instance normalization \cite{ulyanov_instance_2017} or to introduce building blocks from other machine learning domains, such as self-attention \cite{zhang_self-attention_2019}. 

\noindent\textbf{Domain adaptation on RGB-D.} To our knowledge there has not been an RGB-D to RGB-D translation for domain adaptation where both, the image domain and the depth domain, are adapted simultaneously. We have found several applications where depth data is used along RGB data, i.e. depth-depth adaptation, and depth-RGB-adaptation. \cite{atapour-abarghouei_generative_2019} uses domain adaption to reconstruct the depth map without holes from an RGB-D input whose depth map has missing information caused by the depth sensor. \cite{gan_light-weight_2021} and \cite{hossain_efficient_2022} generate depth maps from a given stereo-image input. \cite{hambarde_depth_2020} reconstructs the depth map given only a monocular image and semantic map and \cite{islam_depth_2021} uses RGB-D data during training and RGB data during testing to reconstruct the depth map.

\section{Proposed Method}
\label{sec:proposed_method}

\begin{figure*}[h]
  \centering
  \includegraphics[width=0.9\textwidth]{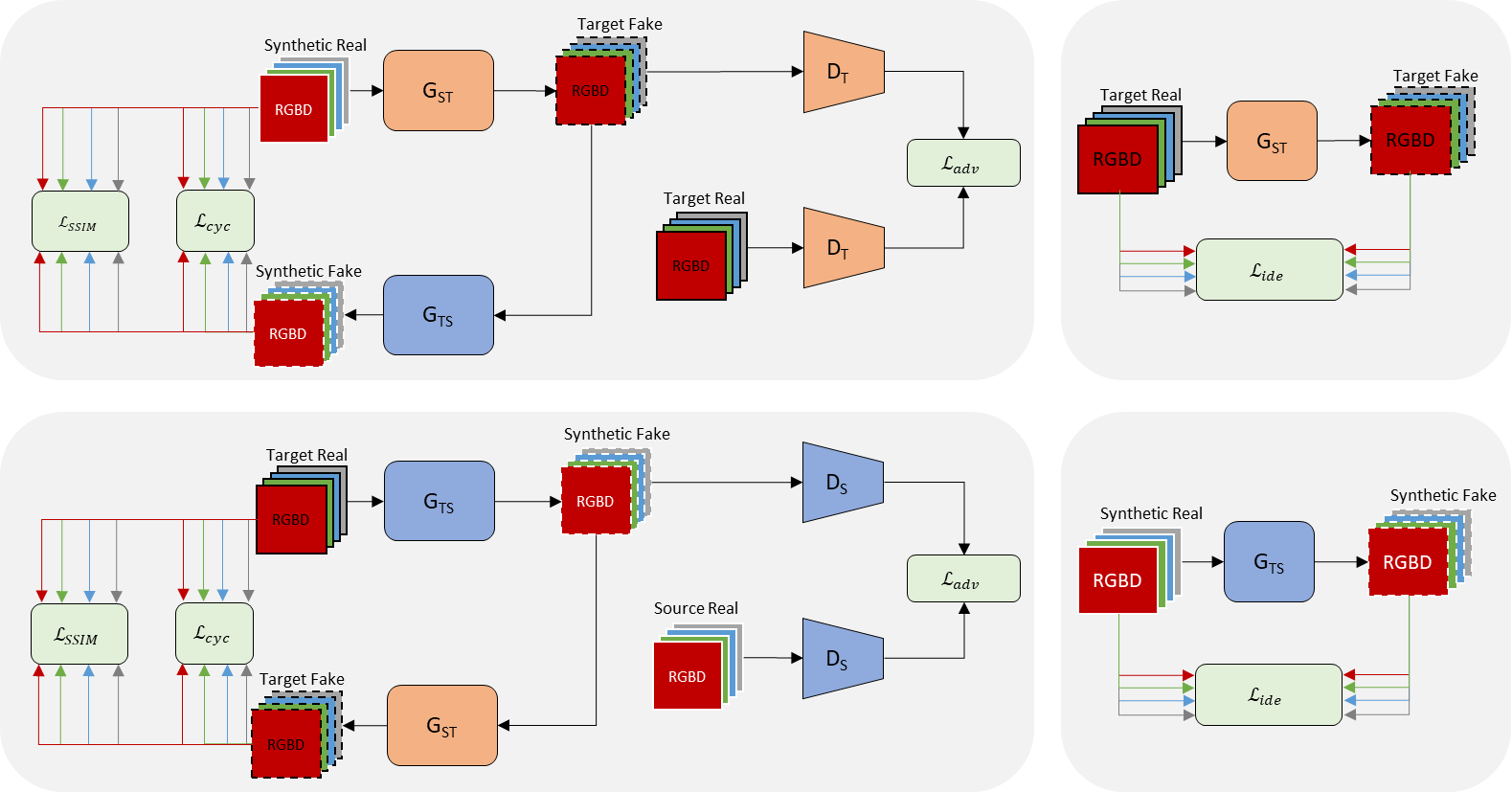}
  \caption{VoloGAN architecture. Our framework consists of four models: two generators and two discriminators. The generators translate an RGB-D image from one domain into the respective other domain. The discriminators predict whether a generated RGB-D is real or fake. We incorporate four loss terms: adversarial loss, a channel-wise cycle-consistency loss, a channel-wise structural similarity loss of cycled image pairs and an identity loss.}
  \label{fig:vologan}
\end{figure*}

We propose VoloGAN, a framework for adversarial domain adaptation. VoloGAN is based on the CycleGAN architecture. The architecture is shown in Fig.\ref{fig:vologan}. The framework implements two generators and two discriminators. The generators $G_{TS}$ and $G_{ST}$ translate the RGB-D images from target domain to synthetic domain and from synthetic domain to target domain respectively. The discriminators $D_{S}$ and $D_{T}$ estimate, whether an input RGB-D image from the synthetic domain or target domain is real or fake.

VoloGAN further modifies its loss functions compared to the CycleGAN. First, for the adversarial loss we calculate the mean squared error (MSE) as proposed by \cite{mao_least_2017} instead of the binary cross entropy (BCE) to improve training stability. The adversarial loss for the different domains of the VoloGAN is, therefore,

\begin{equation}
    \begin{aligned}
        & \mathcal{L}_{adv}\left( G_{ST}, D_T\right) = 
        \mathbb{E}_{t \sim p_{data}(t)}\left[\left( D_T(t)-1\right)^2\right] 
        +
        \mathbb{E}_{s \sim p_{data}(s)}\left[\left(D_T\left(G_{ST}(s)\right)\right)^2\right],\\ 
        & \mathcal{L}_{adv}\left( G_{TS}, D_S\right) =
        \mathbb{E}_{s \sim p_{data}(s)}\left[\left(D_S(s)-1\right)^2\right] 
        +
        \mathbb{E}_{t \sim p_{data}(t)}\left[\left(D_S\left(G_{TS}(t)\right)\right)^2\right] ,
    \end{aligned}
\end{equation}

where $s$ is a sample from the synthetic domain, $t$ is a sample from the target domain.

Inspired by \cite{zhao_loss_2018} we implement a pixel loss $\mathcal{L}_{pix}$ for the cycle consistency loss $\mathcal{L}_{cyc}$ and the identity loss $\mathcal{L}_{ide}$. $\mathcal{L}_{pix}$ switches during training from mean absolute error (MAE) to MSE at the epoch $epoch_{sw}$. It is defined as

\begin{equation}
    \mathcal{L}_{pix}\left(x^p, y^p, x, y\right) = 
    \left\{
        \begin{array}{ll}
            \mathbb{E}_{x \sim p_{data}(x)}\left[||x^p - x||_1\right] + \mathbb{E}_{y \sim p_{data}(y)}\left[||y^p - y||_1\right], & epoch\leq epoch_{sw}  \\
            \mathbb{E}_{x \sim p_{data}(x)}\left[||x^p - x||_2^2\right] + \mathbb{E}_{y \sim p_{data}(y)}\left[||y^p - y||_2^2\right], & epoch > epoch_{sw}
        \end{array}
    \right. .
\end{equation}

Using the pixel loss, the cycle consistency loss becomes

\begin{equation}
    \mathcal{L}_{cyc}\left( G_{ST}, G_{ST}, s, t\right) = \mathcal{L}_{pix}\left(
    G_{TS}\left(G_{ST}(s)\right),
    G_{ST}\left(G_{TS}(t)\right),
    s,
    t
    \right),
\end{equation}

and the identity loss becomes

\begin{equation}
    \mathcal{L}_{ide}\left( G_{ST}, G_{TS}, s, t\right) = \mathcal{L}_{pix}.
    \left(
        G_{TS}(s), G_{ST}(t), s, t
    \right),
\end{equation}

Additionally, we add a structural similarity loss $\mathcal{L}_{SSIM}$ that calculates the similarity of images based on their structural similarity index (SSIM). SSIM compares contrast, luminance, and structure of two images using statistical parameters i.e., the mean $\mu$, the variance $\sigma$, and the covariance $\sigma_{x,y}$ of both images. The SSIM of two images $x$ and $y$ is defined as

\begin{equation}
    SSIM(x,y)=
    \underbrace{\left[ \frac{2\mu_x\mu_y+C_1}{\mu_x^2+\mu_y^2 +C_1}\right]^{\alpha}}_\text{contrast} \cdot
    \underbrace{\left[ \frac{2\sigma_x\sigma_y+C_2}{\sigma_x^2+\sigma_y^2 +C_2}\right]^{\beta}}_\text{luminance} \cdot
    \underbrace{\left[ \frac{\sigma_{x,y}+C_3}{\sigma_x\sigma_y+C_3}\right]^{\gamma}}_\text{structure} ,
\end{equation}

where $\alpha$, $\beta$ and $\gamma$ are hyperparameters to give relative importance to individual terms and $C_1$, $C_2$ and $C_3$ are constants that must be chosen.
The structural similarity loss $\mathcal{L}_{SSIM}$ is calculated on a cycled image and an unaltered image of both domains respectively. It is defined as

\begin{equation}
\mathcal{L}_{SSIM}\left( G_{ST}, G_{TS}, s, t\right) = \left[ 1- SSIM(s,G_{TS}\left(G_{ST}(s)\right))\right] + \left[1-SSIM(t,G_{ST}\left(G_{TS}(t)\right)) \right],
\end{equation}

To suppress the effect of channel pollution caused by the convolution of the early fused RGB-D images, we propose to calculate certain loss functions channel-wise as in \cite{tang_gesturegan_2019}. The respective loss function $\mathcal{L}_{\Lambda}$ of each channel $i$ is calculated independently, scaled by $\lambda_{i}$ and summed up. In contrast to \cite{tang_gesturegan_2019}, we apply it to multiple loss terms so that 

\begin{equation}
\mathcal{L}_{\Lambda,channel}\left( G_{ST}, G_{TS}, s, t\right)= \sum\limits_{i \in \{r,g,b,d \}}\lambda^i \mathcal{L}_\Lambda^i\left( G_{ST}, G_{TS}, s, t\right), with \, \Lambda \in \{cyc, ide,ssim\}.
\end{equation}

We set $\lambda^d=3$ and $\lambda^{\{r,g,b\}}=1$ to compensate for the underrepresented depth channel (three color channels vs. one depth channel). Finally, the loss function of VoloGAN is the weighted sum of all loss terms presented before as

\begin{equation}
\mathcal{L}_{VoloGAN} = \mathcal{L}_{adv} + \lambda_{cyc}\mathcal{L}_{cyc,channel} + \lambda_{ide}\mathcal{L}_{ide,channel} + \lambda_{ssim}\mathcal{L}_{ssim,channel},
\end{equation}

where we set $\lambda_{cyc}=10$, $\lambda_{ide} = 0.5$ and $\lambda_{ssim} = 1$.

\subsection{Generator}

\begin{figure*}[h]
  \centering
  \includegraphics[width=0.9\textwidth]{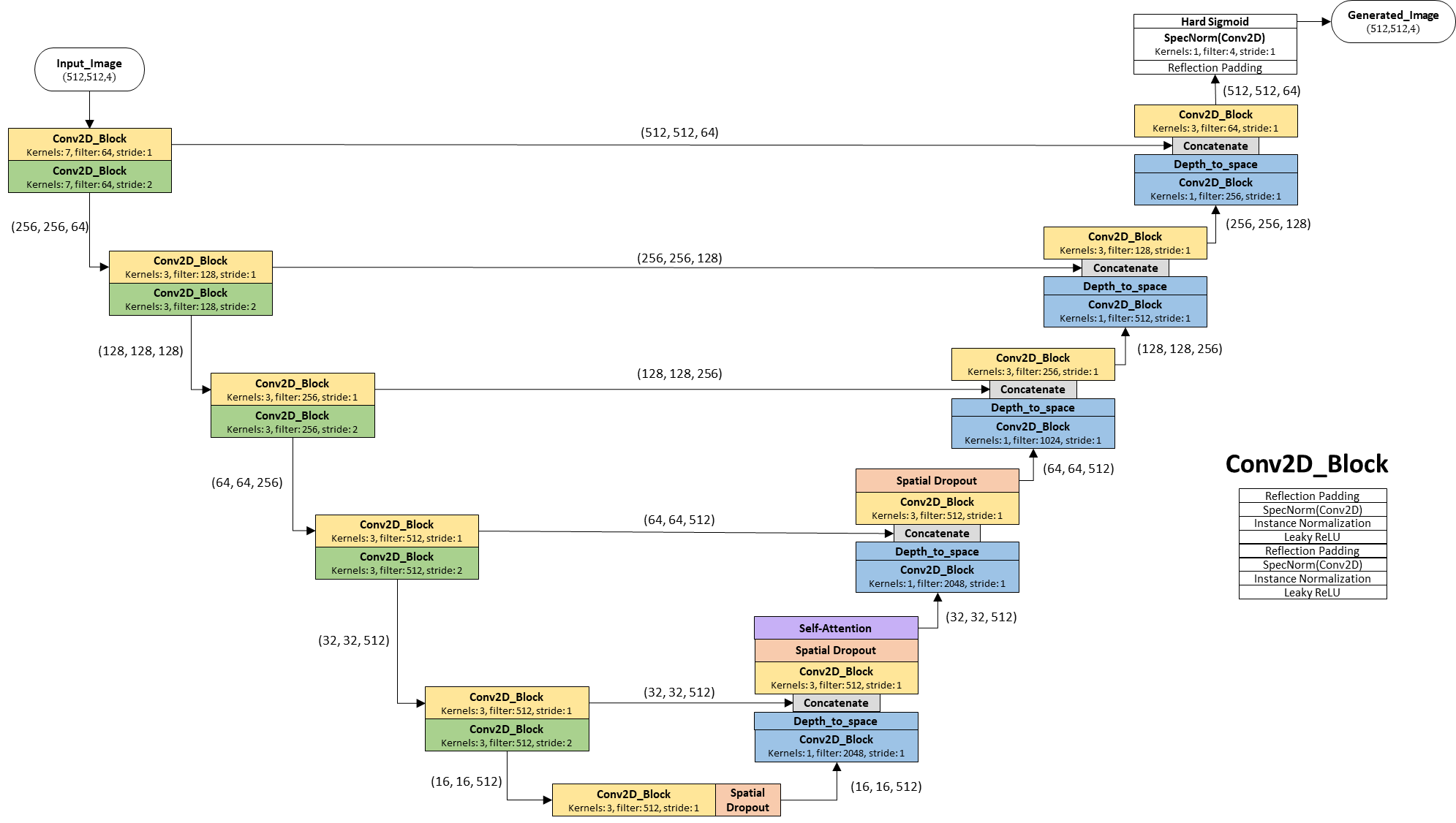}
  \caption{VoloGAN generator. Our generator follows an encoder-decoder architecture with multiple connections between encoder and decoder.}
  \label{fig:generator_model}
\end{figure*}

For our purpose, the generator has to translate one RGB-D image from the synthetic data distribution $p_{synthetic}$ to the generator’s distribution $p_g$ that is as close as possible to the real data distribution $p_{target}$. This means that the input and the output dimensions of the generator need to be equal to the shape of the input image, i.e., 512x512x4. Furthermore, we primarily want to alter local features (e.g. disturbances in the LiDAR scan) keeping the global context (e.g. pose of the person). We selected a U-Net based generator that has been used in segmentation tasks such as \cite{ronneberger_u-net_2015}. Our generator model is depicted in Fig.\ref{fig:generator_model}. 

The architecture consists of six levels. For downsampling, a strided convolution is applied whereas in the upsampling path the depth-to-space transformation is applied. The generator’s encoder-decoder architecture and the skip level connections between them, allows information to flow from the encoder to the decoder which can be interpreted as combining the content from the latent space with the respective locality of that feature. Our model has six different levels. With each level, the spatial width is decreased by a factor of two, while the number of channels is doubled. The conv2d\_block consists of two consecutive convolutions applying spectral normalization \cite{miyato_spectral_2018} and instance normalization \cite{ulyanov_instance_2017}. Exception is the last layer, where the number of channels is kept at 512, to save memory. All convolutions use filters with a kernel size of three, except of the first layer that uses a kernel size of seven to increase the perceptive field while having only a few channels. Due to the extensive padding in the TPU caused by residual and aggregated residual blocks, it was not possible to implement these in our model, since the U-Net itself is already prone to padding due to the concatenations between encoder and decoder. To keep the dimension of feature maps fixed, we apply padding. Specifically, we use reflection padding to retain the distribution of the feature map. We use spatial dropout \cite{tompson_efficient_2015} in the first three stages of the decoder. Self-attention with a trainable gate (see appendix \ref{sec:trainable_selfattention}) is applied in the decoder path on the stage with a spatial width of 32x32. Generally, it is recommended on higher spatial dimensions, but due to performance and memory constraints it could not be applied. We use leaky ReLU activations, since we found it is a good trade-off between model performance and resources required for training. The output of the generator is activated with a hard sigmoid function to limit the output to values between one and zero. In contrast to the standard sigmoid function, the hard sigmoid function can represent the values one and zero by clipping rather than only asymptotically approximating said values. 

In this configuration, the U-Net model has in total 39,390,917 parameters, of which 14,276 are non-trainable due to the spectral normalization layer. For the VoloGAN, this number has to be doubled since two generators are used, resulting in 78,781,834 parameters only for the generators.

\subsection{Discriminator}
\begin{figure*}[h]
  \centering
    \includegraphics[width=0.6\textwidth]{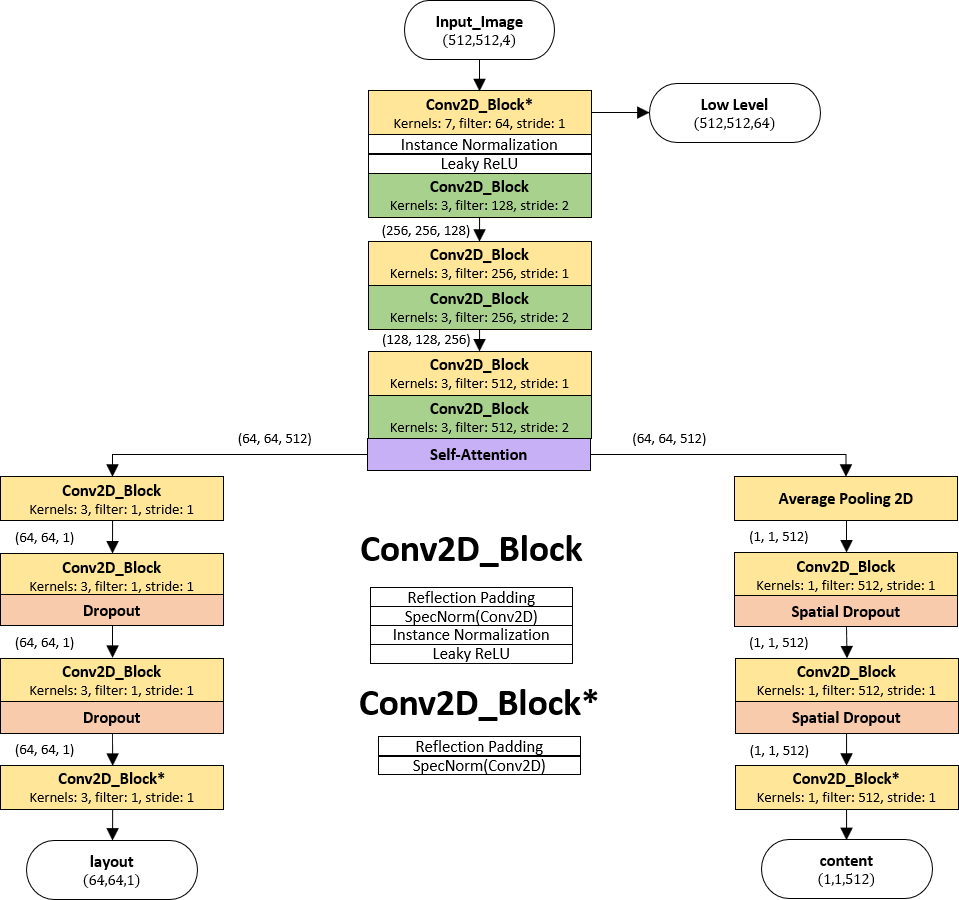}
  \caption{VoloGAN discriminator. Our discriminator has three outputs to evaluate whether an input RGB-D image is real or fake: low level evaluation, layout evaluation and content evaluation. We explicitly encourage the disentanglement between layout and content by a two-branch architecture.}
  \label{fig:critic_model}
\end{figure*}

The discriminator of the VoloGAN is depicted in Fig.\ref{fig:critic_model}. Its implementation is based on \cite{sushko_generating_2021}, but in contrast to the original model, we only discriminate full sized images. 
It has three different outputs: low-level, content and layout. Its encoder has three downsampling stages before it is branched into the content branch and the layout branch. The conv2d\_block uses reflection padding, spectral normalization, instance normalization and is activated by the leaky ReLU non-linearity. In contrast to the generator, it has only one consecutive convolution with a kernel size of 3. As in the generator, the first convolution has a 7x7 kernel to increase the field of perception. Gated self-attention is applied after the encoder. Dropout is applied in the layout and in the content branch. Since the content branch has multiple channels, spatial dropout is applied, whereas in the content branch only one channel is present, hence standard dropout is applied. Otherwise, the entire feature map would be dropped resulting in zero. 
The adversarial loss $\mathcal{L}_{adv}^{D_{oneshot}}$ of the discriminator is the weighted sum of the low-level loss $\mathcal{L}_{adv,lowlevel}$, the layout loss $\mathcal{L}_{adv,layout}$ and the content loss $\mathcal{L}_{adv,content}$ described by

\begin{equation}
\mathcal{L}_{adv}^{D_{oneshot}} = 2 \mathcal{L}_{adv,lowlevel} + \mathcal{L}_{adv,layout} + \mathcal{L}_{adv,content}
\end{equation}

The low-level loss is multiplied by two to compensate for the two branches. The original SIV-GAN discriminator implements a diversity regularization, that encourages a disentangled latent space of the noise vector. Since the VoloGAN follows the CycleGAN framework and not the vanilla GAN framework, this additional loss is not suitable and therefore left out.
In addition, the SIV-GAN discriminator has been designed to be able to learn from a single training sample, which is beneficial since VoloGAN is trained with only few samples.

In this configuration, our discriminator has in total 9,385,686 parameters, of which 5,640 are non-trainable due to the spectral normalization layer. For the VoloGAN, this number must be doubled since two discriminators are used, resulting in 18,771,372 parameters only for the discriminators.

\subsection{Model Training}
\label{sec:model_training}

Our model is trained on Google’s TPUv3 that has 8 cores and a maximum count of floating-point operations per second of 420TeraFLOPS. We followed a synchronous data parallelism scheme, where the model is stored in each core, trained on a different data batch, and the results are aggregated on a parameter server. We trained on 2048 RGB-D files from each domain with a global batch size of eight, hence each core trained on a local batch size of one. We used an 80/20 train/test split and trained for 80 epochs. We linearly increased the learning rate the first 10 epochs and used cosinusodial learning rate decay for the remaining epochs. We applied a target learning rate of 0.0002 for the generator and 0.0001 for the discriminator. Since the discriminator has fewer parameters, we decreased the learning rate to ensure the discriminator does not outperform the generator. Furthermore, we used NADAM optimizer with $\beta_1=0.5$ and $\beta_2=0.99$ for the generator and SGD with momentum of 0.9 for the discriminator. 
Its common practice to use the He-initializer when using ReLU activations. Many authors also recommend using the He-initializer when applying leaky ReLU activations. We observed that the positive slope for values smaller than zero in the leaky ReLU activation is not considered in Tensorflow’s implementation, as it was originally intended by \cite{he_delving_2015}. Therefore, the initialization values of a layer’s $l$ parameters $W^{[l]}$  with $n^{[l]}$  values in that layer is drawn from a uniform distribution $\mathcal{U}$.

\begin{equation}
W^{[l]} \sim \mathcal{U}\left(-\sqrt{\frac{6}{n^{[l]}+n^{[l+1]}}},  \sqrt{\frac{6}{n^{[l]}+n^{[l+1]}}}\right)
\end{equation}

To save the need for large random-access memory (RAM) we streamed the dataset on demand into the RAM. We applied data augmentation to virtually increase the size of the dataset. Particularly, we randomly flipped images from left to right and randomly shifted the person depicted in the image. The depth has been updated accordingly.

\begin{figure}[h]
\centering
\begin{subfigure}{.25\textwidth}
  \centering
  \includegraphics[width=0.9\linewidth]{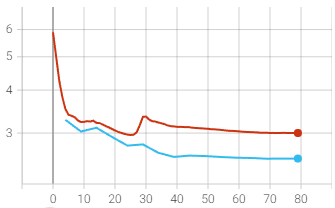}
  \caption{Gen1: Adversarial Loss}
  \label{fig:gen_s_t_adv}
\end{subfigure}%
\begin{subfigure}{.25\textwidth}
  \centering
  \includegraphics[width=0.9\linewidth]{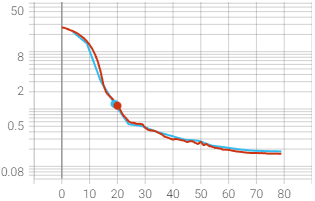}
  \caption{Gen1: Cycle Loss}
  \label{fig:gen_s_t_cyc}
\end{subfigure}%
\begin{subfigure}{.25\textwidth}
  \centering
  \includegraphics[width=0.9\linewidth]{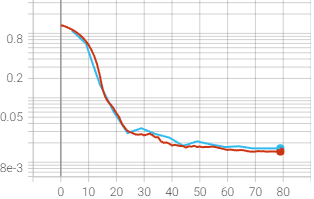}
  \caption{Gen1: Identity Loss}
  \label{fig:gen_s_t_ide}
\end{subfigure}%
\begin{subfigure}{.25\textwidth}
  \centering
  \includegraphics[width=0.9\linewidth]{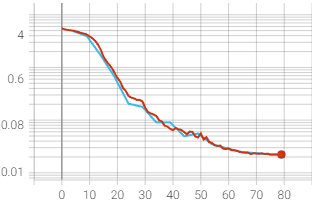}
  \caption{Gen1: SSIM Loss}
  \label{fig:gen_s_t_ssim}
\end{subfigure}%
\\
\begin{subfigure}{.25\textwidth}
  \centering
  \includegraphics[width=0.9\linewidth]{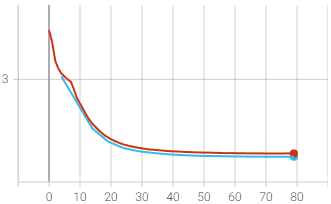}
  \caption{Gen2: Adversarial Loss}
  \label{fig:gen_t_s_adv}
\end{subfigure}%
\begin{subfigure}{.25\textwidth}
  \centering
  \includegraphics[width=0.9\linewidth]{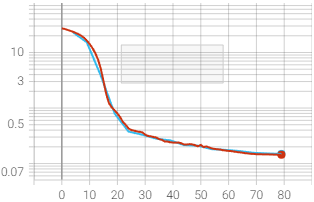}
  \caption{Gen2: Cycle Loss}
  \label{fig:gen_t_s_cyc}
\end{subfigure}%
\begin{subfigure}{.25\textwidth}
  \centering
  \includegraphics[width=0.9\linewidth]{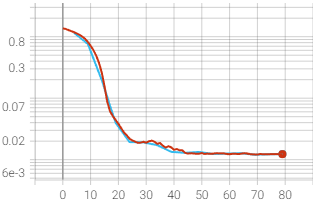}
  \caption{Gen2: Identity Loss}
  \label{fig:gen_t_s_ide}
\end{subfigure}%
\begin{subfigure}{.25\textwidth}
  \centering
  \includegraphics[width=0.9\linewidth]{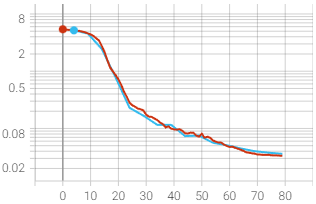}
  \caption{Gen2: SSIM Loss}
  \label{fig:gen_t_s_ssim}
\end{subfigure}%
\caption{Training metrics. Train metric recorded every epoch, test metric recorded every 5 epochs.}
\label{fig:training_metrics}
\end{figure}

The training progress of 80 epochs of both generators is shown in Fig. \ref{fig:training_metrics}. These metrics depict the individual loss terms of each generator model (i.e., the adversarial loss , the cycle loss, the identity loss, and the structural similarity loss).

In summary, all training and test losses are converging. It has been observed that the model could have been trained for more epochs, indicated by the fact that all loss terms are further decreasing. Due to the timing limitations in the usage of the training hardware, it was not possible to train more than 9 hours. There are no signs of overfitting in the learning curve. It must be considered, though, that due to the low variance of the dataset, similar samples are present in both, the training and the test set, which could be biasing the result of the test set.
\section{Results and Discussion}
\label{sec:results_and_discussion}

\begin{figure}[h]
\centering
\begin{subfigure}{.5\textwidth}
  \centering
  \includegraphics[width=0.95\linewidth]{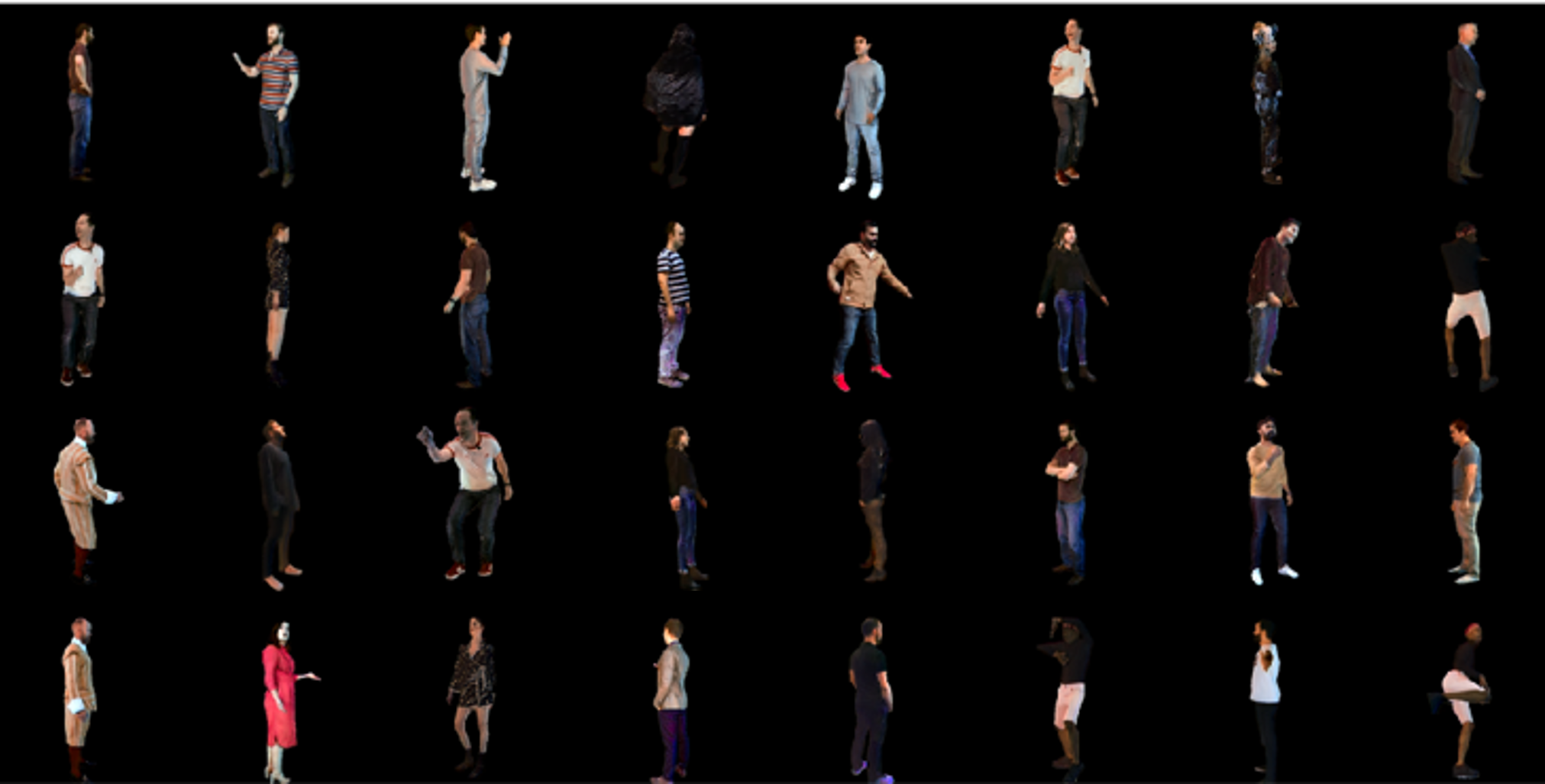}
  \caption{}
  \label{fig:multiple_rgb}
\end{subfigure}%
\begin{subfigure}{.5\textwidth}
  \centering
  \includegraphics[width=0.95\linewidth]{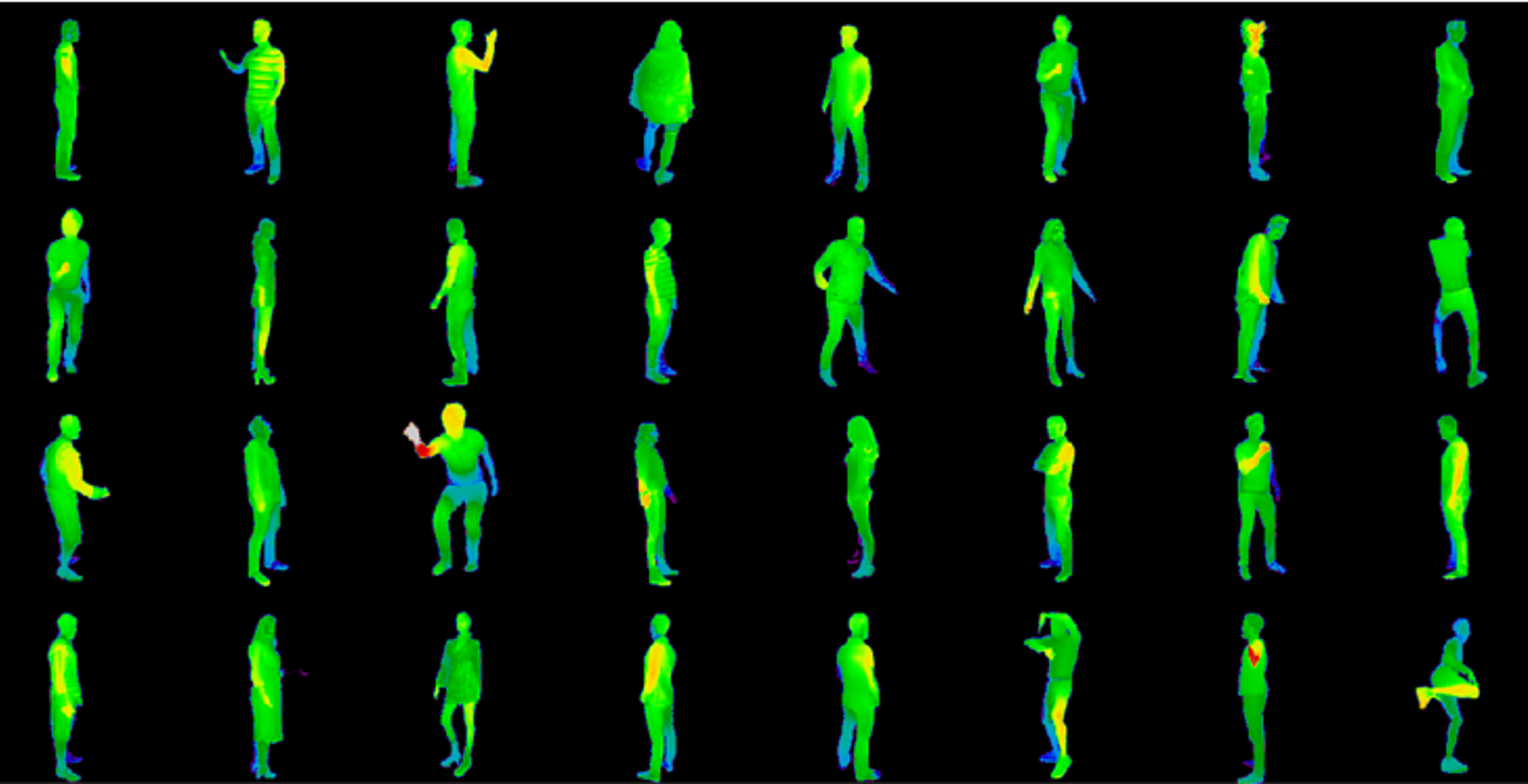}
  \caption{}
  \label{fig:multiple_depth}
\end{subfigure}
\caption{Generated images into the target domain from the synthetic domain. (a) RGB channel, (b) Depth channel. }
\label{fig:multiple_RGB_depth}
\end{figure}

In this section we present and discuss our results. In a first step, multiple generated images are qualitatively observed. Fig.\ref{fig:multiple_RGB_depth} shows images that have been generated from the synthetic domain into the target domain. Qualitatively, the model has learned to generate images of persons. It has learned that there is a black background without objects. The global context is preserved indicated by the fact that the pose of the persons as well as the type of cloths have not been changed. Hence, the geometric structure has been generated successfully. 
The color space seems to be more saturated, realistic and some colors of the clothing have changed. 
The associated depth channels are presented in Fig.\ref{fig:multiple_depth}. The distance of a point in the depth map is color encoded. Red points are closer and purple points are farther away. It can be observed that there is almost no channel pollution in the depth which would be indicated by features that could only be visible in the color space (e.g., shirt with colored pattern would have different depths in the depth map). Hence, the channel-wise loss with extra weight in the depth channel is effective. Finally, some arms and leg are slightly thinner than before. This is a consequence of the non-ideal segmentation mask of the images in the target domain, some extremities have been cut, and the generator must reconstruct them.

\begin{figure}[h]
\centering
\begin{subfigure}{.9\textwidth}
  \centering
  \includegraphics[width=1\linewidth]{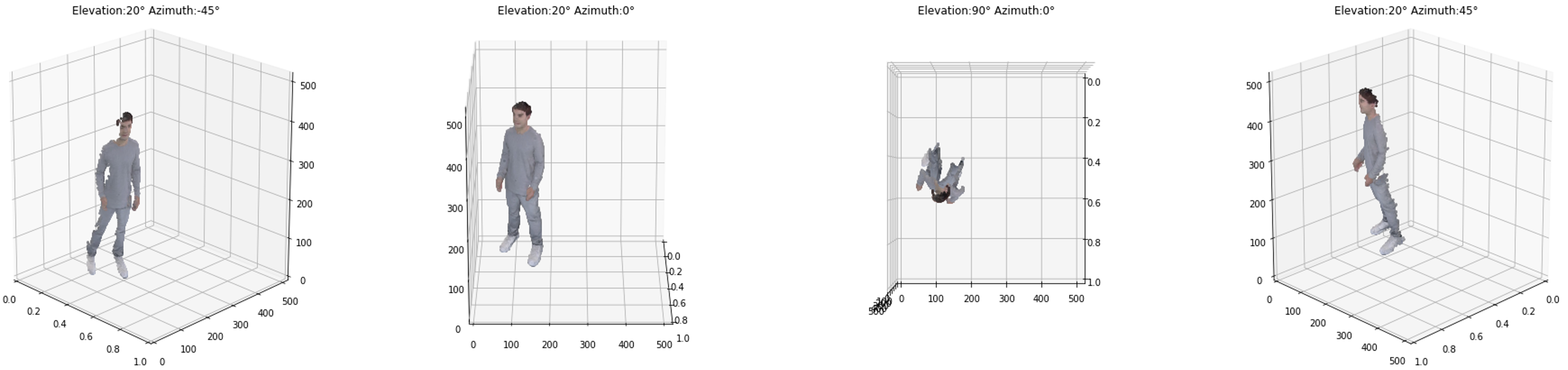}
  \caption{}
  \label{fig:3d_pointcloud_input}
\end{subfigure}%
\\
\begin{subfigure}{.9\textwidth}
  \centering
  \includegraphics[width=1\linewidth]{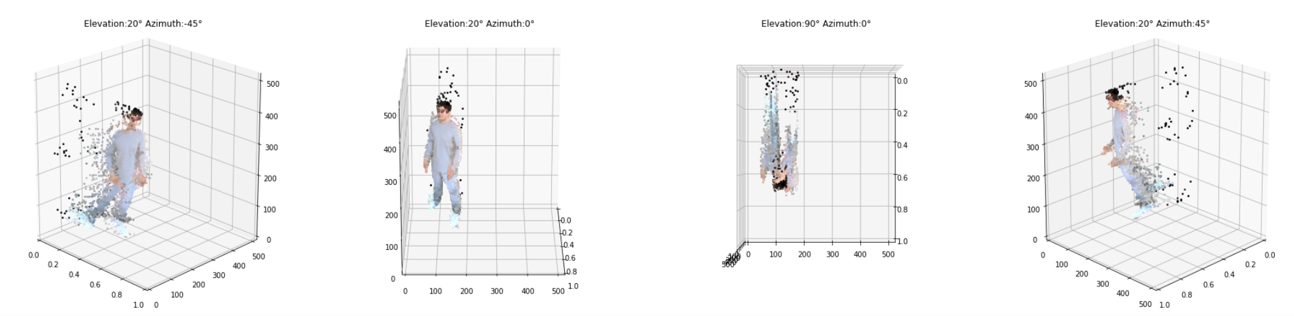}
  \caption{}
  \label{fig:3d_pointcloud_generated}
\end{subfigure}
\caption{3D point cloud. (a) before training, (b) after training 80 epochs. }
\label{fig:3d_point_cloud}
\end{figure}

Fig.\ref{fig:3d_point_cloud} combines the color and the depth space into a 3D point cloud. Fig.\ref{fig:3d_pointcloud_input} shows a point cloud taken from the synthetic distribution $p_{synthetic}$ that serves as input into the target generator $G_{ST}$. Fig.\ref{fig:3d_pointcloud_generated} depicts the generated output. Due to performance reasons during training, only every $2^{nd}$ point is plotted and zeros indicating the background have been ignored.

It is observable that the position and shape of the person has not been altered. Furthermore, a tail of scattered points has been generated behind the person, which is a main characteristic of the target domain caused by non-ideal segmentation masks and the physical characteristics of the LiDAR scanner. Appendix \ref{sec:dataset} provides more details on the dataset.

Since we are mainly interested in the performance of the target generator, we will evaluate its results in more detail.

\begin{figure}[h]
  \centering
  \includegraphics[width=0.7\textwidth]{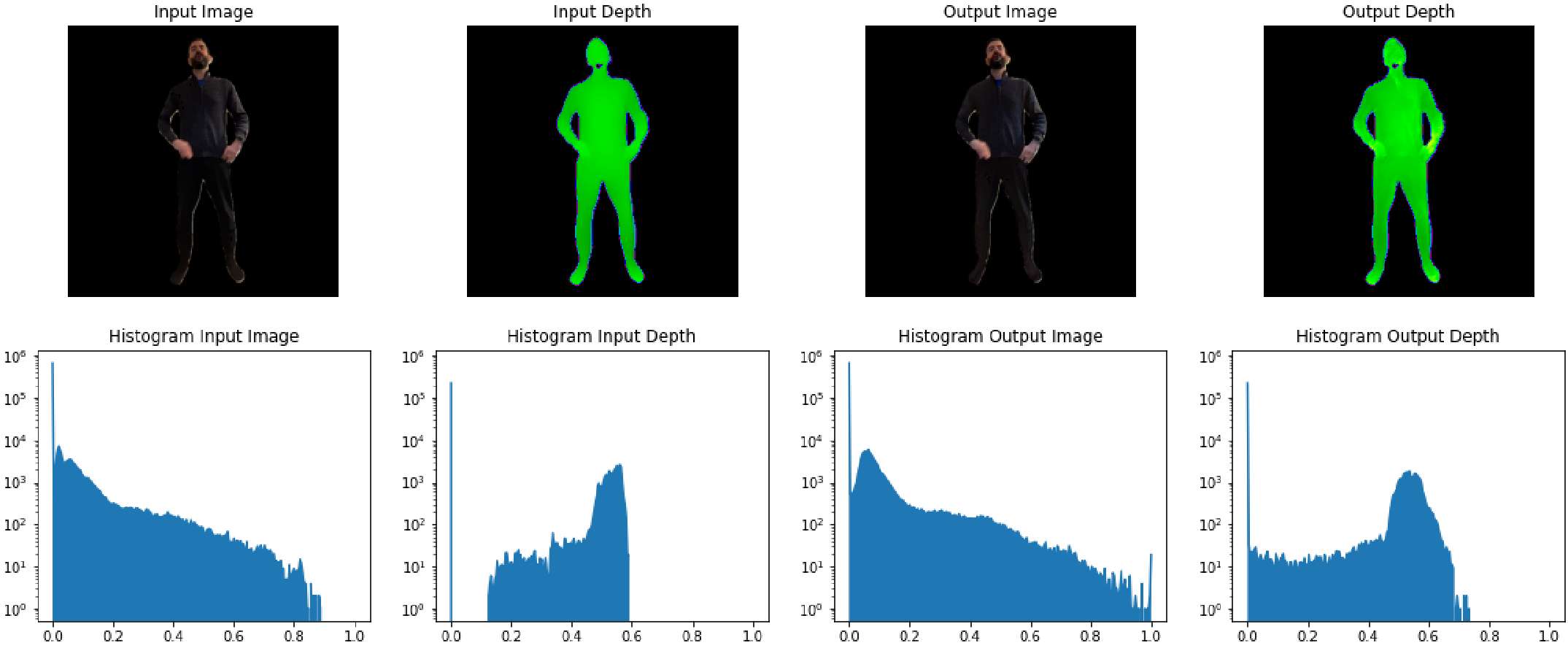}
  \caption{Target generator output for an input image taken from the target distribution}
  \label{fig:generator_identity}
\end{figure}

The identity mapping is depicted in Fig. \ref{fig:generator_identity}. An image from the target domain is passed through the target generator that is supposed to generate images of the target domain. Ideally, the image should not be altered at all since it is already from said domain. A high similarity is achieved. The image space appears to be similar, with slight signs of an altered color space and nearly no signs of channel pollution in the depth map. For the histograms, the peaks are at comparable positions. As for the noise in the output depth histogram, one must consider the logarithmic scale.

\begin{figure}[h]
  \centering
  \includegraphics[width=0.7\textwidth]{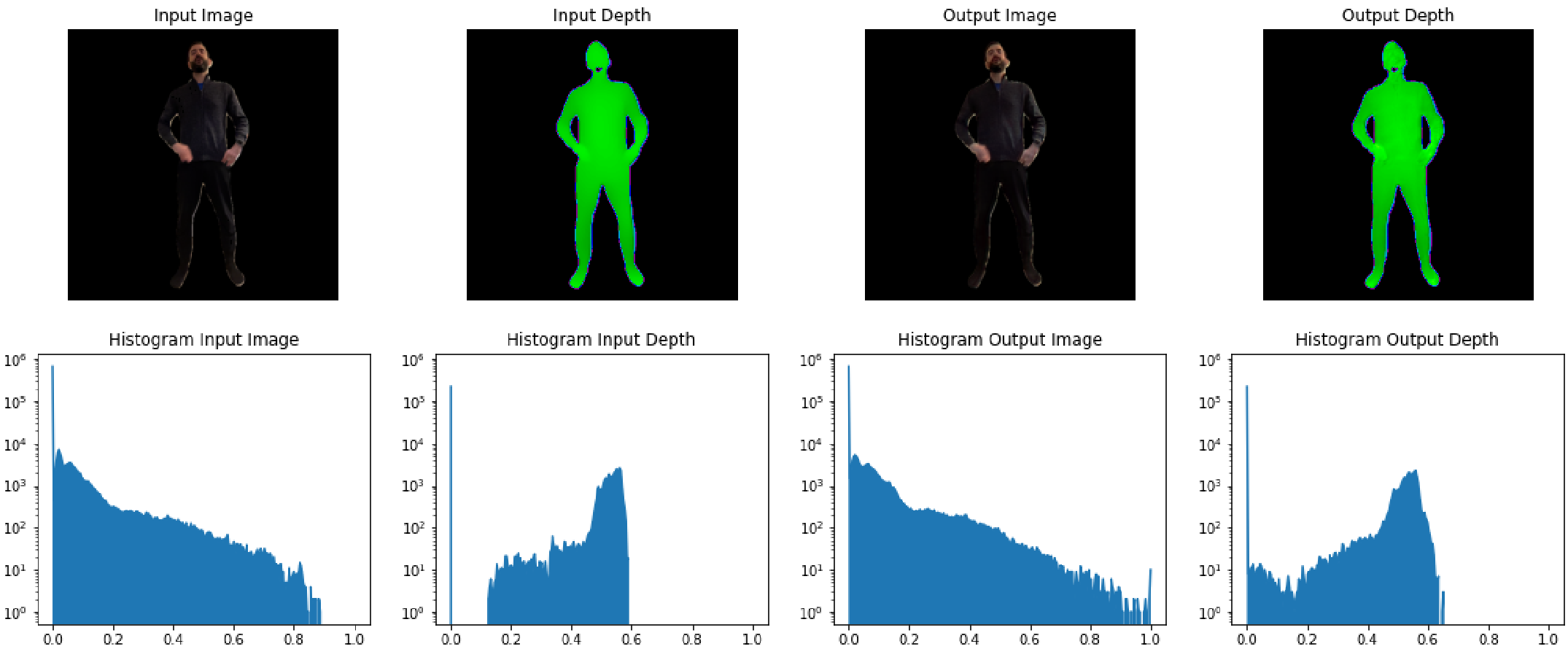}
  \caption{Cycled input image taken from the synthetic domain}
  \label{fig:generator_cycle}
\end{figure}

Fig.\ref{fig:generator_cycle} depicts the cycled output image for an input image that has been first passed through the target generator and then passed through the synthetic generator back into its original domain. Ideally, the cycled image would appear exactly like the initial image. The resulting output looks similar to the initial input. In comparison with the identity mapping, the color is better preserved, and the depth maps show less channel pollution. This is due to the high importance given to the cycle consistency loss $\mathcal{L}_{cyc,channel}$ by setting $\lambda_{cyc}$ to 10, while in case of the identity loss $\mathcal{L}_{ide,channel}$ the weighting factor $\lambda_{ide}$ is set to 0.5

\begin{figure}[h]
  \centering
  \includegraphics[width=0.7\textwidth]{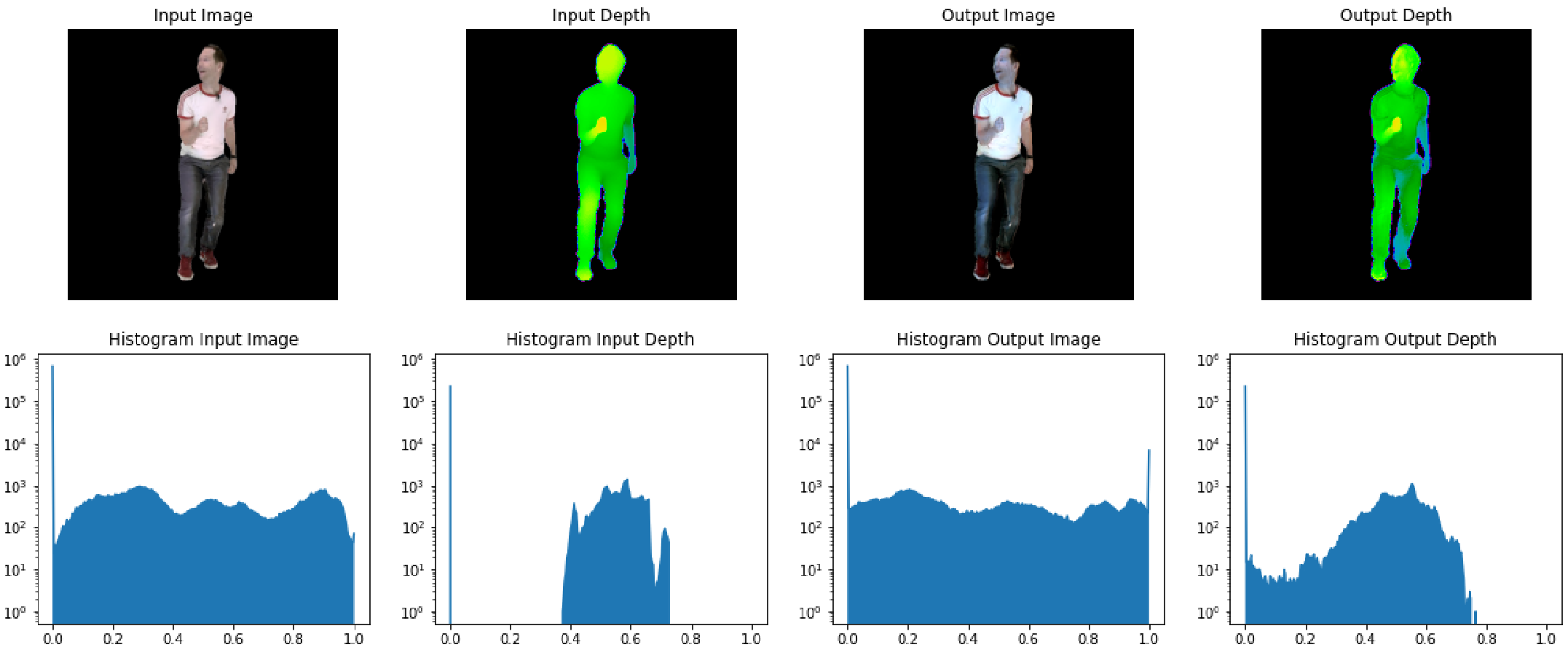}
  \caption{Target generator output for an input image taken from the synthetic distribution}
  \label{fig:generator_target}
\end{figure}

Finally, the images translated from the synthetic domain into the target domain using the adversarial domain adaptation network are presented. These images are of special interest since they determine the performance of the adversarial domain adaptation task. The evaluation of the ideal result can only be done qualitatively, since there is no ground truth on how an image would look like in the other domain and other typical metrics such as the FID score \cite{heusel_gans_2018} cannot be applied on RGB-D inputs. Fig.\ref{fig:generator_target} depicts the translation from synthetic to target domain. The colors are preserved but appear slightly washed out. This could be an effect of the higher relevance given to the depth map. It might be improved if the model is trained for more epochs. Slight indications of channel pollution are visible. The bluish spots in the depth map on the legs, indicate that the person appears to be tilted, which is a feature that is more distinct in the target domain as in the synthetic domain. Also, the model has learned to preserve the distance of the person in the depth map. This is also indicated by the histogram of the depth map, where the peak of the input’s and the output’s distributions are roughly at the same position. Furthermore, it can be observed in the histogram, that values have been inserted between 0 and the main block. This is a distinct feature of the target domain, caused by non-ideal segmentation masks and the physical characteristics of the LiDAR scanner. 

\begin{figure}[h]
\centering
\begin{subfigure}{.45\textwidth}
  \centering
  \includegraphics[width=0.9\linewidth]{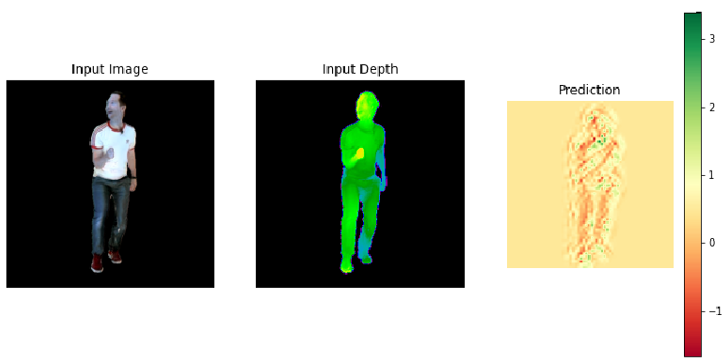}
  \caption{}
  \label{fig:critic_fake}
\end{subfigure}%
\begin{subfigure}{.45\textwidth}
  \centering
  \includegraphics[width=0.9\linewidth]{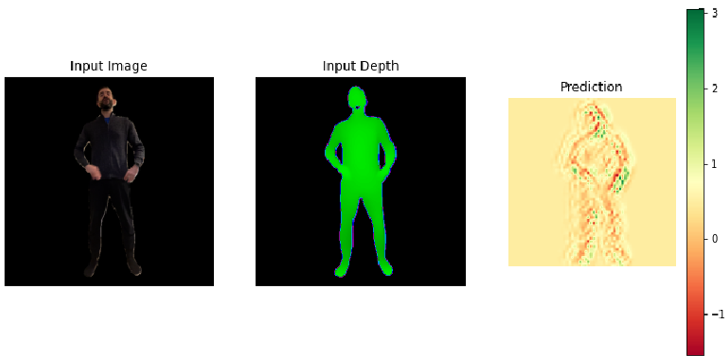}
  \caption{}
  \label{fig:critic_real}
\end{subfigure}
\caption{Layout output target discriminator (a) fake input, (b) real input.}
\label{fig:critic_output}
\end{figure}

Finally, we show the layout output of the target discriminator in Fig.\ref{fig:critic_output}.

Zero predicts a value to be fake, while one predicts a real value. Values higher than one over emphasize a certain true classified value and negative values below zero overemphasize fake values respectively. The values 0 and 1 are defined by the ground truth label used during training. When comparing two different predictions, it must be considered that the color bar is set dynamically depending on the values present in the feature map and does not represent an absolute value. In the figure above, most of the values (especially the background) are predicted to be near 0.5, with some exceptions on the person’s body. Hence, the discriminator can’t tell whether an image is real or fake, which is the objective of the adversarial training. The generator generates images so well that the discriminator can’t tell whether it is a fake or not. It is important to note that the values are not exactly 0.5 so the discriminator can still provide valuable feedback to the generators training.

\begin{figure}[h]
\centering
\begin{subfigure}{.45\textwidth}
  \centering
  \includegraphics[width=0.9\linewidth]{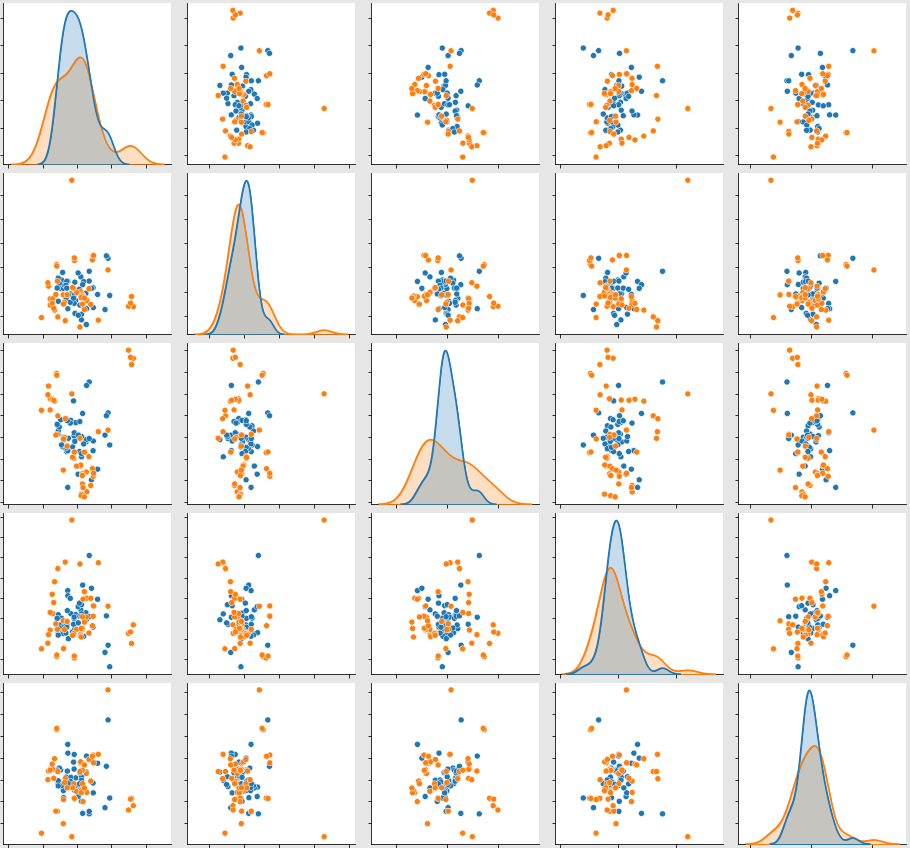}
  \caption{}
  \label{fig:pca_before}
\end{subfigure}%
\begin{subfigure}{.45\textwidth}
  \centering
  \includegraphics[width=0.9\linewidth]{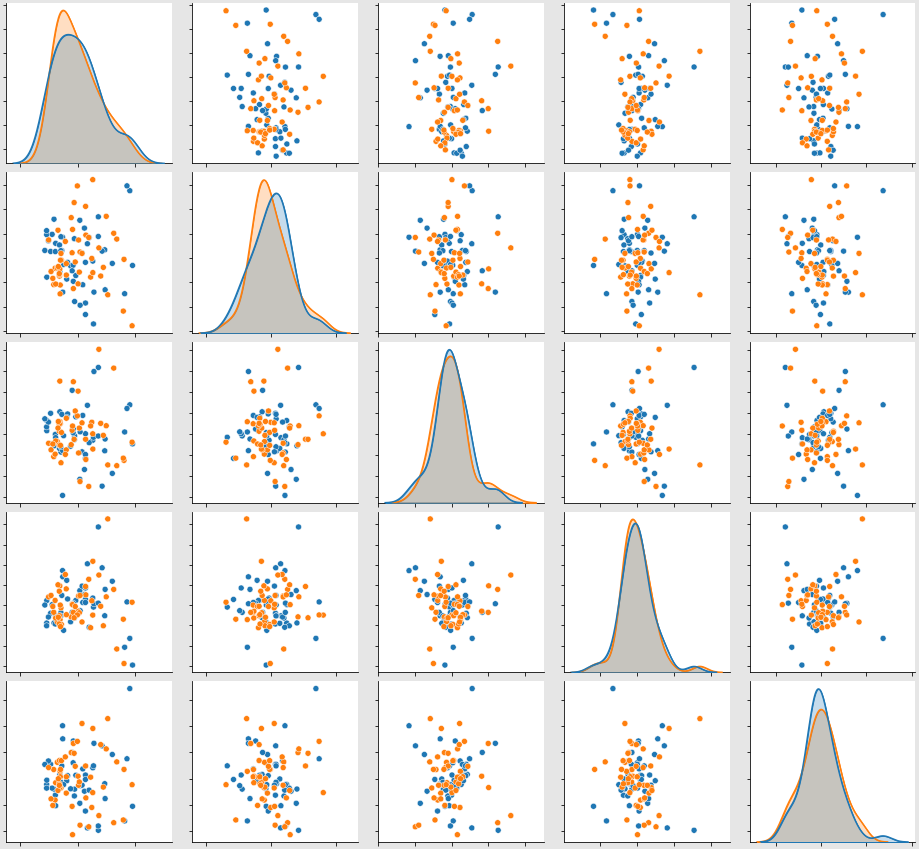}
  \caption{}
  \label{fig:pca_after}
\end{subfigure}
\caption{Principal component analysis of the five principal components from 50 samples of each domain. Orange: generated images of the target domain.  Blue: real images of the target domain. (a) before training, (b) after training 80 epochs.}
\label{fig:pca}
\end{figure}

A principal component analysis is performed on generated and real RGB-D images of the target domain. In that way, the similarity of two distributions can be evaluated, based on their most significant features. 
Fig.\ref{fig:pca} shows a PCA analysis of the five principal components of 50 real (blue dots) and 50 generated (orange dots) images of the target domain before and after training. Before training, the generated images have high variance, meaning that images do not share similar principal components. The real images are less distributed. After 80 epochs of training, the generated distribution approaches the real distribution closely and the values are more distributed.

\section{Conclusion}
\label{sec:conclusion}
It has been demonstrated that VoloGAN is suitable for the adversarial domain adaptation task. RGB-D images could be successfully translated from one domain into the respective other domain with acceptable performance. VoloGAN is capable to reconstruct depth maps with the characteristics of a consumer depth sensor.
The complexity of the adversarial domain adaptation task, the usage of the CycleGAN-based framework and the large spatial width of the images contributed to the final complexity of the VoloGAN model. Due to its high number of trainable parameters, the usage of hardware accelerators was inevitable. Furthermore, due to the constraints with respect to available training hardware, the most practical solution was to use Kaggle’s TPUv3, to be able to train a model in feasible time, considering the need for design iterations, as well. The usage of a TPU highly constraints the model architecture. For that reason, architectural compromises, like the avoidance of residual layers, had to be made. It has been shown that by carefully selecting other design parameters (e.g. activations, optimizer, initializer and advances custom layers), acceptable results could be achieved. 

Due to the trade-offs made in the model architecture, training the model became more difficult. CycleGANs are known to be unstable during training, due to their adversarial training objective of multiple models (i.e., generator and discriminator). Measures to stabilize the training, such as spectral normalization, instance normalization and self-attention have been implemented. In that way, it was possible to train the VoloGAN consisting of a U-Net based generator and SIV-GAN based discriminator. It has been shown that despite of the different modalities of the color and the depth space, good results can be achieved when concatenating both modalities into a single RGB-D image (early fusion), supported by a channel-wise loss function. Furthermore, the discriminator has shown superior performance with its multi-branch output architecture that disentangles content features from layout features.

\newpage
\bibliographystyle{unsrt}  
\bibliography{references}  

\newpage
\section*{Author biographies}
\begin{multicols}{3}
    \begin{wrapfigure}{l}{25mm} 
        \includegraphics[width=1in,height=1.25in,clip,keepaspectratio]{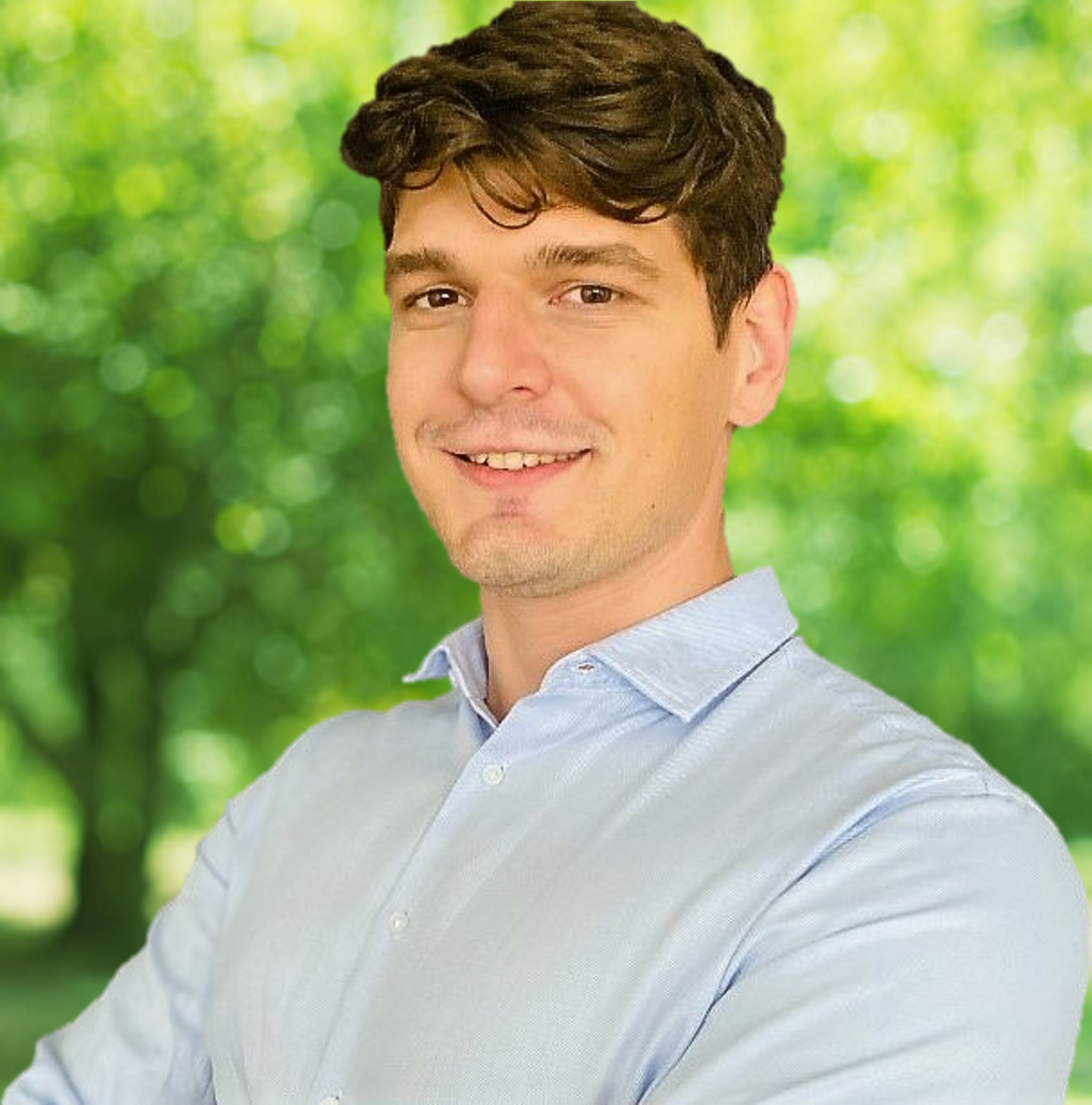}
    \end{wrapfigure}\par
    \noindent \textbf{Sascha Kirch} is a doctoral student at UNED (National University for Distance Education), Spain. His research focuses on self-supervised multi-modal generative deep learning. In parallel to his studies, Sascha works as technical lead in the development of automotive radar electronics at Bosch, Germany. He received his master’s degree in Electronic Systems for Communication and Information from UNED, Spain. He received his bachelor’s degree in electrical engineering from the Cooperative State University Baden-Wuerttemberg (DHBW), Germany. Sascha is member of IEEE’s honor society Eta Kappa Nu as part of the chapter Nu Alpha.\\
    \vfill\null
    \columnbreak
    
    \setlength\intextsep{0pt} 
    \begin{wrapfigure}{l}{25mm} 
        \includegraphics[width=1in,height=1.25in,clip,keepaspectratio]{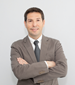}
    \end{wrapfigure}\par
    \noindent \textbf{Sergio Martín} is Associate Professor at UNED (National University for Distance Education, Spain). He is PhD by the Electrical and Computer Engineering Department of the Industrial Engineering School of UNED. He is Computer Engineer in Distributed Applications and Systems by the Carlos III University of Madrid. He teaches subjects related to microelectronics and digital electronics since 2007 in the Industrial Engineering School of UNED. He has participated since 2002 in national and international research projects related to mobile devices, ambient intelligence, and location-based technologies as well as in projects related to "e-learning", virtual and remote labs, and new technologies applied to distance education. He has published more than 200 papers both in international journals and conferences. He is IEEE senior member.\\

    \vfill\null
    \columnbreak
    
    \setlength\intextsep{0pt} 
    \begin{wrapfigure}{l}{25mm} 
        \includegraphics[width=1in,height=1.25in,clip,keepaspectratio]{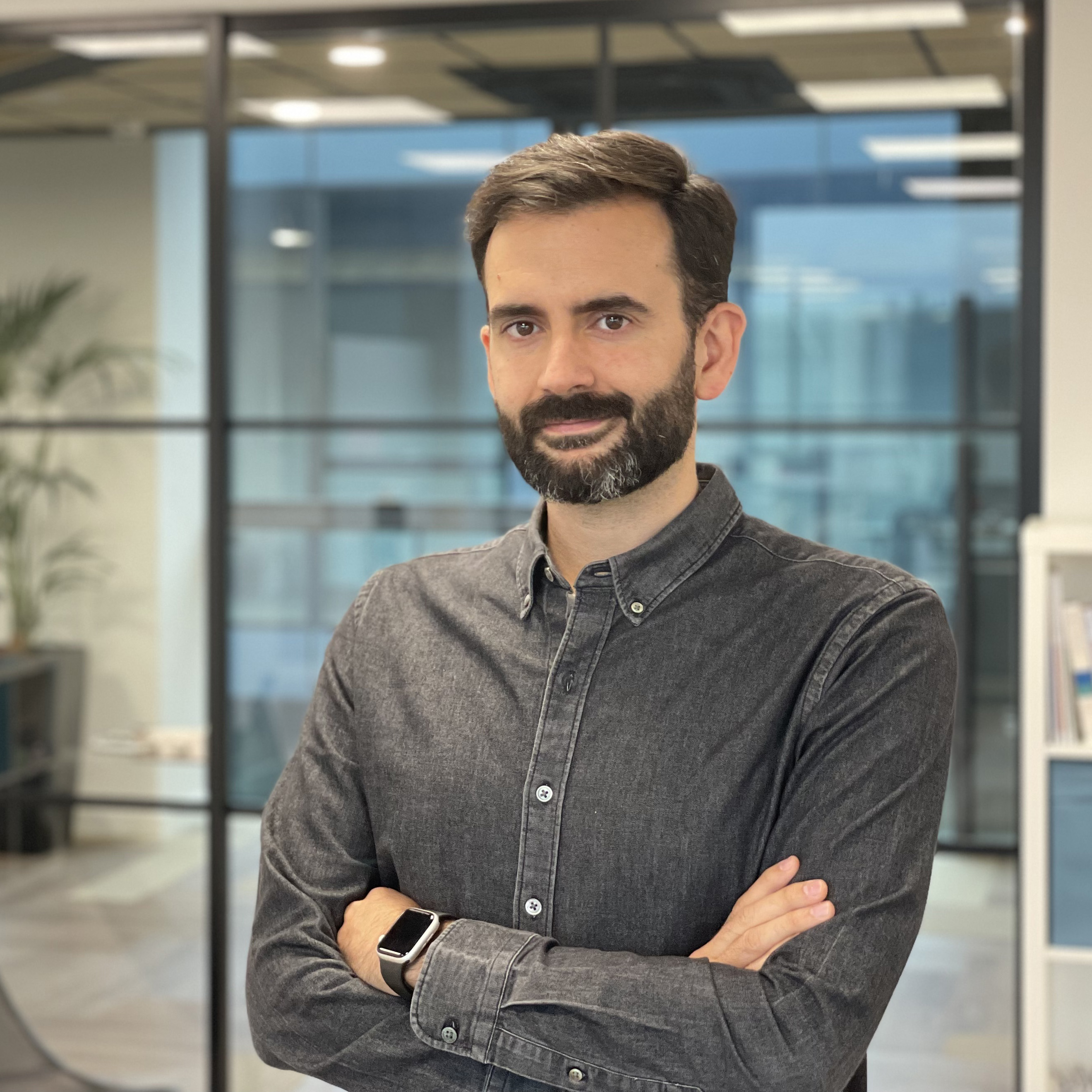}
    \end{wrapfigure}\par
    \noindent \textbf{Rafael Pagés} is Co-Founder and CEO of Volograms, a startup bringing 3D reconstruction technologies to everyone. He received the Telecommunications Engineering degree (Integrated BSc-MSc accredited by ABET) in 2010, and PhD in Communication Technologies and Systems degree in 2016, both from Technical University of Madrid (UPM), in Spain. Rafael was member of the Image Processing Group at UPM and did his post-doctoral research at Trinity College Dublin. His research interests include 3D reconstruction, volumetric video, and computer vision.\\
    
\end{multicols}

\newpage
\appendix
\section{Appendix}
\subsection{Dataset}
\label{sec:dataset}

\begin{figure}[h]
\centering
\begin{subfigure}{.45\textwidth}
  \centering
  \includegraphics[width=0.9\linewidth]{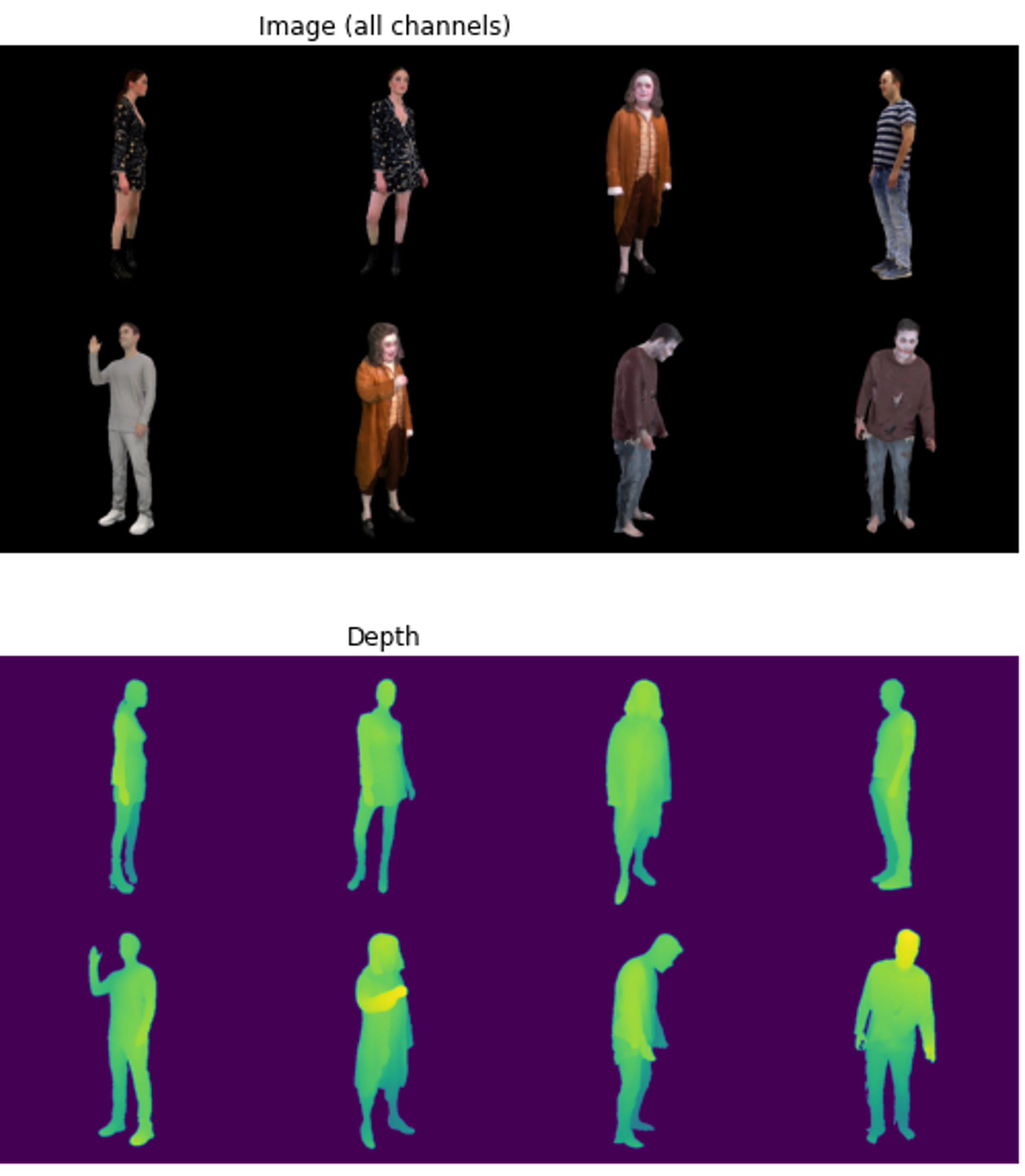}
  \caption{}
  \label{fig:dataset_synthetic}
\end{subfigure}%
\begin{subfigure}{.45\textwidth}
  \centering
  \includegraphics[width=0.9\linewidth]{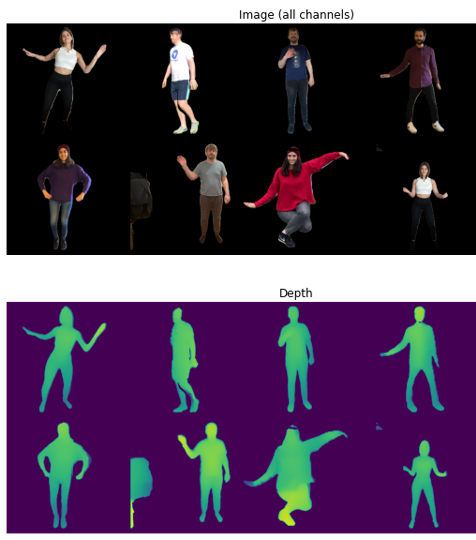}
  \caption{}
  \label{fig:dataset_target}
\end{subfigure}
\caption{Training data examples. (a) Synthetic domain, (b) target domain.}
\label{fig:dataset}
\end{figure}

The dataset used to train the VoloGAN contains 2048 samples from the “synthetic domain”, which are images that are rendered from previously generated 3D models using photogrammetry and 2048 samples from the “target domain” which have been captured using a mobile phone equipped with a LiDAR scanner. Fig. \ref{fig:dataset} shows some examples from both domains. 

A sample contains a person centered in the RGB image in front of a black background and its corresponding depth map. The height and width of the images are 512x512 pixels. The color space of both domains is provided in RGB that is encoded with 8bit (uint8) per channel. Hence, values range from 0 to 255, while the background is zero in all channels. The depth space is provided in float32. The person in the depth space is centered around zero, while the background has the value of minus one. The centered depth is calculated by subtracting the real depth from the distance to the object. That means, the smaller the centered depth, the further the real depth. 
Samples from the synthetic domain are rendered from existing 3D models. To generate samples from the target domain, persons are extracted using a segmentation mask. 

\begin{figure}[h]
\centering 
\begin{subfigure}{.9\textwidth}
  \centering
  \includegraphics[width=1\linewidth]{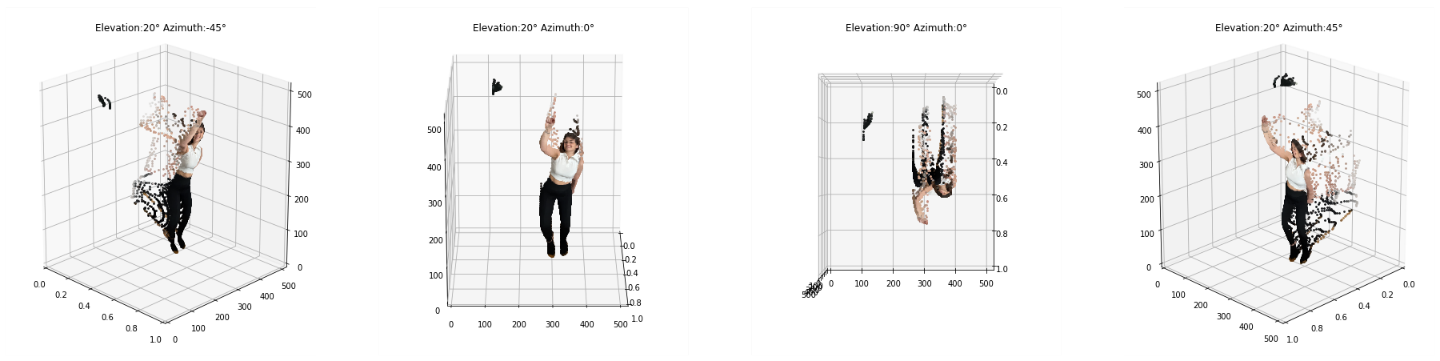}
  \caption{}
  \label{fig:dataset_pointcloud_target}
\end{subfigure}%
\\
\begin{subfigure}{.9\textwidth}
  \centering
  \includegraphics[width=1\linewidth]{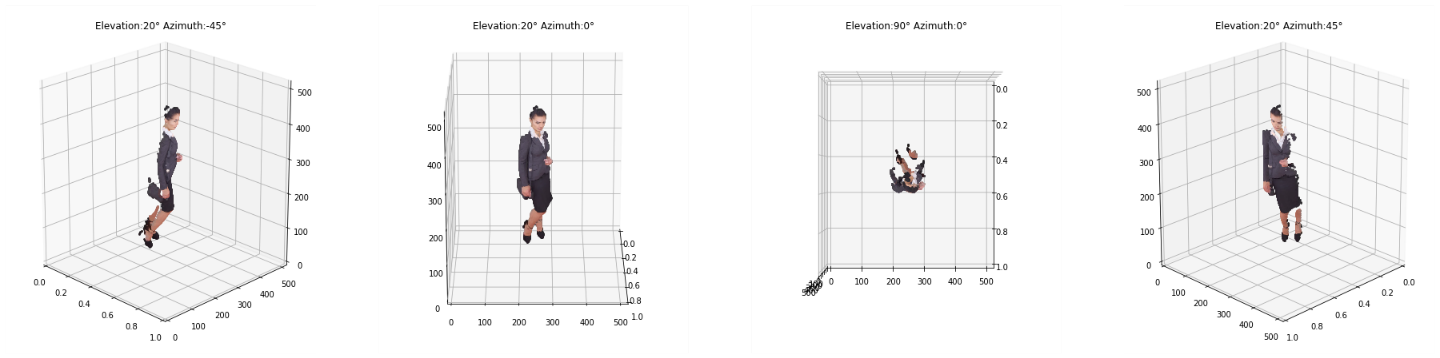}
  \caption{}
  \label{fig:dataset_pointcloud_synthtetic}
\end{subfigure}
\caption{Point cloud comparison of target and synthetic domain samples. (a) target domain, (b) synthetic domain}
\label{fig:dataset_3d_point_cloud}
\end{figure}

Fig.\ref{fig:dataset_3d_point_cloud} visualizes the color and depth information using a colored point cloud in 3D space with different azimuth and elevation angles.  

Two additional characteristics can be extracted from the point cloud. First, images from the target domain have a tail of points in border regions. Second, persons in both domains are slightly tilted. In the target domain, this effect is caused by the angular error due to the position of the LiDAR scanner. In the synthetic domain, this effect is artificially added by randomly moving the viewport during rendering the images.

We apply min-max scaling to the data to stabilize training and to scale both modalities to the same range of values. Each channel of the RGB image is represented in uint8, hence might have values from 0 to 255. Hence, the scaled RGB image is defined as:

\begin{equation}
RGB_{scaled}=\frac{RGB-min(uint8)}{max(uint8)-min(uint8)}.    
\end{equation}

From the depth data, it is known that the minimum value is -1, which presents the background and the maximum value that is 1 due to clipping. Hence, the scaled depth is defined as:

\begin{equation}
Depth_{scaled}=\frac{Depth-min(Depth)}{max(Depth)-min(Depth)}   
\end{equation}

\subsection{Architectural Details}
\label{sec:architectural_details}

\subsubsection{Gated Self-Attention}
\label{sec:trainable_selfattention}
\begin{figure}[h]
  \centering
  \includegraphics[width=0.9\textwidth]{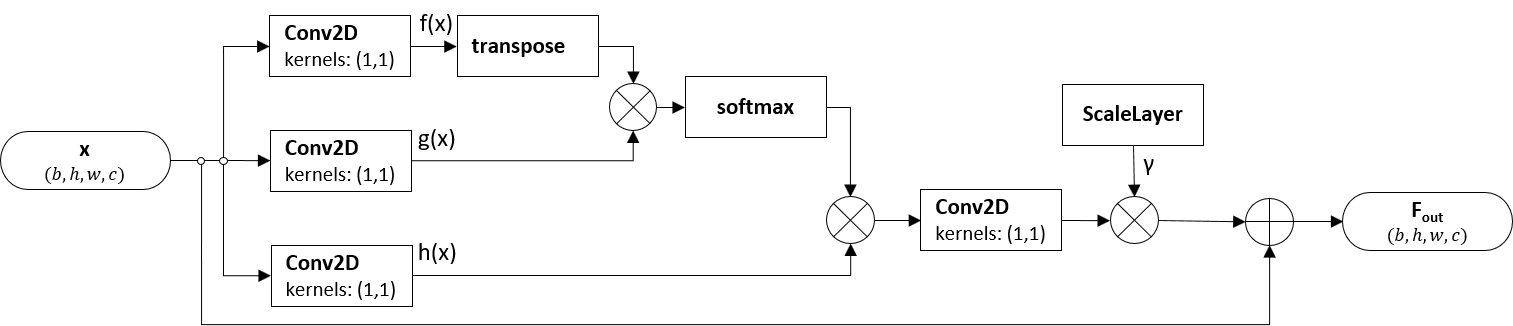}
  \caption{VoloGAN's gated self-attention block.}
  \label{fig:vologans_learnable_attention_block}
\end{figure}

Due to the small spatial width of filters in a convolutional network, the perceptive field is rather small focusing only on the local context of a feature map. The global context is considered by multiple consecutive layers. Self-attention is a mechanism used to consider the global context that has been proven to be very effective in natural language processing \cite{vaswani_attention_2017}, grid-like data \cite{wang_non-local_2018} and even in generative adversarial networks \cite{zhang_self-attention_2019}. Since the computational requirements scale quadratically with the number of pixels, we deployed self-attention in layers with small spatial width.
We adopted the self-attention mechanism of \cite{zhang_self-attention_2019} but implemented it in a residual path that is gated by a learnable scalar $\gamma$. Fig. \ref{fig:vologans_learnable_attention_block} depicts VoloGAN's gated self-attention block.

The 1x1 convolutions represent learnable weight matrices. The symbol $\otimes$ denotes an element-wise multiplication. The output feature map $F_{out}$ is the sum of the self-attention feature maps $o$ multiplied by a learnable scalar $\gamma$ and the convolutional feature map $x$, hence can be defined as:

\begin{equation}
F_{out} = \gamma \left( softmax\left(f(x)\otimes g(x)^T\right)\otimes h(x) \right) + x
\end{equation}

\subsubsection{Reflection Padding}
\label{sec:reflection_padding}
When performing a convolutional operation, the output feature map has a smaller spatial dimension than the input feature map. Since in an image-to-image translation task the input and the output dimensions of the model must be equal (i.e., 512x512x4 pixels), the convolutional layers must implement padding to maintain the spatial dimension of the feature map. Padding increases the input’s spatial dimension by filling with certain values in such a way that the output feature map has the same spatial dimension as the input feature map before padding. 
Many machine learning models, and so the baseline models used for the VoloGAN, implement zero padding, where the image is padded with zeros. In Zero padding, the distribution of values of the feature maps might be heavily affected by inserting zeros to a non-zero feature map. This is especially harmful in adversarial domain adaption, where the overall goal is to generate images of a certain distribution that is as close as possible to the target distribution. 

\begin{figure}[h]
  \centering
  \includegraphics[width=0.6\textwidth]{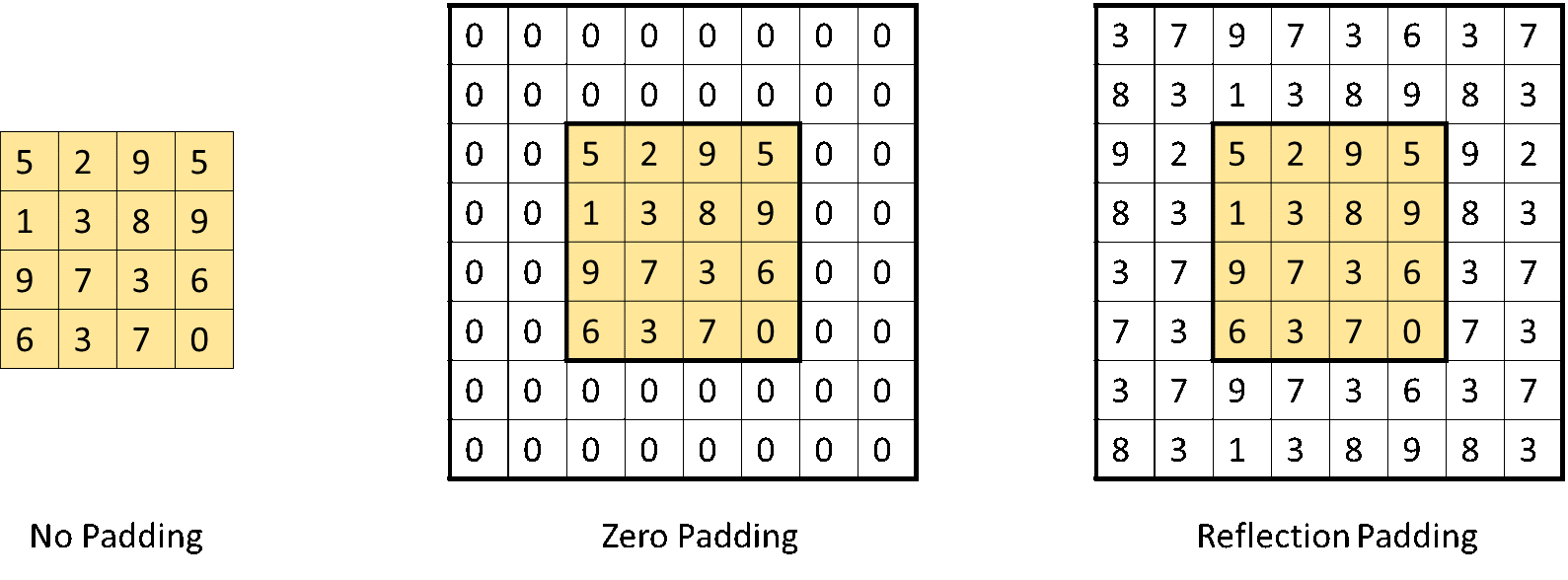}
  \caption{Comparison of zero padding and reflection padding.}
  \label{fig:padding}
\end{figure}

Another approach is using reflection padding, where the value used for padding is determined by the input feature map. Fig. \ref{fig:padding} shows a comparison of both approaches. In reflection padding, the value used for padding is determined by the input feature map. It uses values from within the feature map, i.e., reflects the values across the outer value of the feature map. This approach interferes less with the overall distribution of the values. On the downside, reflection padding is computationally more expensive compared to zero padding. 

\subsubsection{Spatial Dropout}
\label{sec:spatial_dropout}

\begin{figure}[h]
  \centering
  \includegraphics[width=0.6\textwidth]{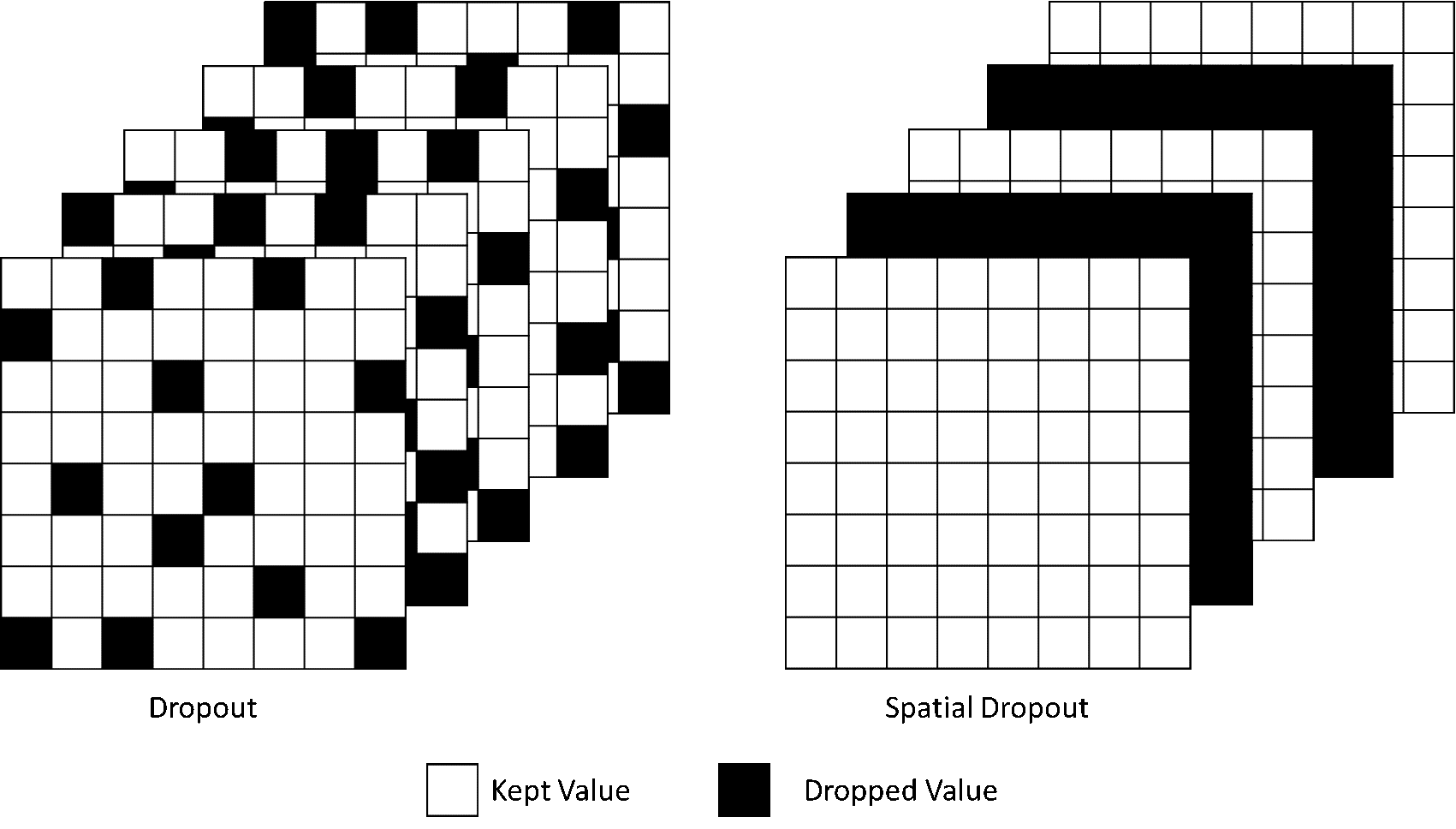}
  \caption{Comparison between dropout and spatial dropout in CNNs.}
  \label{fig:dropout}
\end{figure}

A commonly used technique to prevent large models from overfitting is dropout introduced by \cite{JMLR:v15:srivastava14a}. During training, connections between two layers are randomly and temporarily dropped i.e., set to zero. This reduces the complexity of the model and prevents single parameters from forming stronger connections than others. Both reduce overfitting. Dropout has been introduced in neural networks and has been successfully applied to CNNs as well. However, standard dropout is not that effective in CNNs, especially when combined with normalization techniques since the covariance of nodes cannot be avoided due to the weight sharing of the convolutional filter. An adaption of dropout is spatial dropout \cite{tompson_efficient_2015}, where instead of dropping single values, entire channels are dropped. By dropping a channel instead of single values, not only the performance of CNNs can be improved but it also speeds up training. We applied a dropout rate of 0.2. Fig. \ref{fig:dropout} shows a comparison between standard drop out and spatial dropout in CNNs.

\subsubsection{Up and Down Sampling}
\label{sec:up_down_sampleing}

Upsampling and downsampling are main features the generator and the discriminator architectures presented in the previous section. For both, several approaches can be used.

In the VoloGAN generator, downsampling is used in the encoder to decrease the spatial dimensions while increasing the number of channels to extract features from an image. The generator implements a strided convolutions with a stride of 2. We preferred strided convolutions over pooling since convolutions have learnable parameters and can, therefore, learn the optimum downsampling, while the pooling layer has no learnable parameters and either takes the maximum value or the average of the receptive field, hence information is lost. 

Upsampling in VoloGAN's generator is used to increase the spatial dimension while decreasing the depth of the feature map. The image is reconstructed from the latent space, learning how to generate an image based on the features of the image in the latent space. Two approaches are used in many decoder architectures: transpose convolutions and upsampling layers using an interpolation function (like bilinear, or k-means) followed by convolution. Transpose convolution inserts zero values into the image, while in upsampling intermediate pixels are interpolated considering surrounding pixels. The advantage of transposed convolutions is that by introducing zero values, the computation is fast and no hyperparameter must be selected. Caused by the induced zero values, the upsampled images tend to have a checker-board pattern that is undesirable. These artefacts can be avoided when using an interpolating upsampling layer followed by a convolution to adapt the depth of the feature map. This approach has two disadvantages. First, it is computationally more expensive, since the convolution is applied after the upsampling, hence on a larger image \cite{sugawara_checkerboard_2019}. Second, and more importantly, since values are interpolated, no new information is added to the resulting feature map \cite{wojna_devil_2019}.

\begin{figure}[h]
\centering
\begin{subfigure}{.9\textwidth}
  \centering
  \includegraphics[width=1\linewidth]{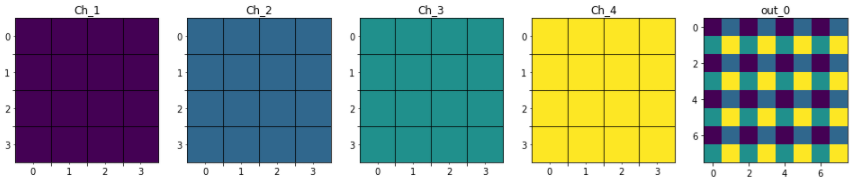}
  \caption{}
  \label{fig:depth_to_space_1}
\end{subfigure}%
\\
\begin{subfigure}{.9\textwidth}
  \centering
  \includegraphics[width=1\linewidth]{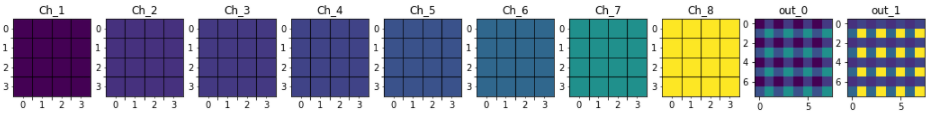}
  \caption{}
  \label{fig:depth_to_space_2}
\end{subfigure}
\\
\begin{subfigure}{.9\textwidth}
  \centering
  \includegraphics[width=1\linewidth]{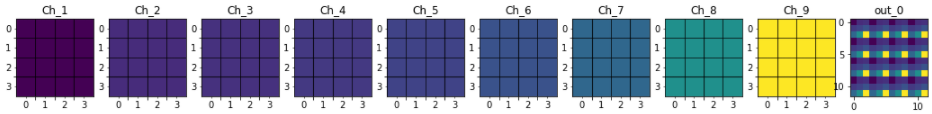}
  \caption{}
  \label{fig:depth_to_space_3}
\end{subfigure}
\caption{Depth to space transformation (a) r=2, channels=4, (b) r=2, channels=8, (c) r=3, channels=9}
\label{fig:depth_to_space}
\end{figure}

The upsampling method used in VoloGAN's generator is the depth-to-space transformation, that is based on the sub-pixel convolution \cite{shi_real-time_2016}. This approach is used successfully in super-resolution applications. Instead of inducing additional values into the feature map, the feature dimension is unraveled to increase the spatial dimension in a non-overlapping manner. The dimensions of the input $(batch,height,width,channel)$ change with the transformation to $(batch,r\cdot height,r\cdot width,\frac{channel}{r^2} )$, where $r$ is the block size. To match the desired output dimensions, this transformation is preceded by a convolution. This upsampling paradigm does not induce additional values; hence, it is more efficient from an information usage point of view. In addition, this transformation is computationally faster than the other two upsampling methods presented before. The difference between the depth-to-space transformation and the sub-pixel convolution is that the latter unravels all channels $(r=\sqrt{channel})$, while the depth-to-space transformation can configure a block size $r$. In the VoloGAN, the block size $r$ is two. Fig.\ref{fig:depth_to_space} depicts the depth-to space transformation using different block sizes and number of channels.

\end{document}